\newif\ifarxiv
\arxivtrue 

\ifarxiv
    \documentclass[]{fairmeta} 
    \usepackage{algorithm}
    \usepackage{amsmath, amssymb}

\else 
    \documentclass{article} 
    \usepackage{icml2025,times}

    \input{math_commands.tex}
    
    \usepackage{icml2025}

    \usepackage[capitalize,noabbrev]{cleveref}

\fi

\usepackage{microtype}
\usepackage{graphicx}
\usepackage{booktabs} 
\usepackage{pgfplots}
\usepackage{hyperref}

\usepackage{amsmath}
\usepackage{amssymb}
\usepackage{mathtools}
\usepackage{amsthm}
\usepackage{tabularx}
\usepackage{booktabs} 
\usepackage{caption} 
\usepackage{xcolor}
\usepackage{natbib}
\usepackage{adjustbox}
\usepackage{xcolor}
\usepackage{caption}
\usepackage{subcaption}
\usepackage{wrapfig}
\usepackage{multirow} 
\usepackage{listings}
\usepackage{algpseudocode}
\usepackage{enumitem}
\PassOptionsToPackage{table}{xcolor}\usepackage{tcolorbox}
\usepackage{sidecap}

\theoremstyle{plain}

\theoremstyle{definition}

\theoremstyle{remark}

\def\myparagraph#1{\noindent\textbf{#1}\hspace{0mm}}

\newcommand{\new}[1]{{\color{black}#1}}
\usepackage[textsize=tiny]{todonotes}

\ifarxiv
    \author[1, 2, 3]{Rim Assouel}
    \author[1]{Florian Bordes}
    \author[1]{Pietro Astolfi}
    \author[1]{Michal Drozdzal}
    \author[1, 2, 4, 5]{Adriana Romero-Soriano}
    
    \affiliation[1]{FAIR at Meta - Montreal}
    \affiliation[2]{Mila}
    \affiliation[3]{Universit\'{e} de Montr\'{e}al}
    \affiliation[4]{McGill University}
    \affiliation[5]{Canada CIFAR AI chair}
    
    \abstract{
    Recent advances in vision language models (VLM) have been driven by contrastive models such as CLIP, which learn to associate visual information with their corresponding text descriptions. However, these models have limitations in understanding complex compositional scenes involving multiple objects and their spatial relationships. To address these challenges, we propose a novel approach that diverges from commonly used strategies, which rely on the design of hard-negative augmentations. Instead, our work focuses on integrating inductive biases into pre-trained CLIP-like models to improve their compositional understanding without using any additional hard-negatives. To that end, we introduce a binding module that connects a scene graph, derived from a text description, with a slot-structured image representation, facilitating a structured similarity assessment between the two modalities. We also leverage relationships as text-conditioned visual constraints, thereby capturing the intricate interactions between objects and their contextual relationships more effectively. Our resulting model not only enhances the performance of CLIP-based models in multi-object compositional understanding but also paves the way towards more accurate and sample-efficient image-text matching of complex scenes.\looseness-1
    }
    
    \date{Feb. 5, 2025}
    \correspondence{\email{assouelr@mila.quebec}}
    
\fi

\ifarxiv
    \title{Object-centric Binding in Contrastive Language-Image Pretraining}
\else
    \icmltitlerunning{Submission and Formatting Instructions for ICML 2025}
\fi

\begin{document}

\ifarxiv
    \maketitle
\fi

\section{Introduction}

Recent advancements in multi-modal representation learning have primarily been enabled by the introduction of CLIP~\citep{radford2021learning}. CLIP learns aligned image-text representations from Internet-scale data. Despite its success, CLIP exhibits limitations in understanding complex scenes composed of multiple objects~\citep{kamath2023whatsupvisionlanguagemodels, yuksekgonul2023visionlanguagemodelsbehavelike,doveh2023teachingstructuredvisionlanguageconcepts,paiss2023teaching}. For instance, while capable of recognizing individual objects, CLIP struggles with interpreting spatial relationships among objects in the scene(\textit{e.g.}, ``the cat is to the left of the mat'' \textit{vs.} ``the cat is to the right of the mat'') and adequately associating objects with their corresponding attributes (\textit{e.g.}, ``a red square and a blue circle'' \textit{vs.} ``a blue square and a red circle''). The process of acquiring this compositional understanding of the world is known as the \emph{binding problem} in the literature, and may be decomposed into \emph{segregation}, \emph{representation}, and \emph{composition} problems~\citep{greff2020bindingproblemartificialneural}.\looseness-1

Efforts to improve the compositional understanding of CLIP-like models have largely relied on leveraging \textit{hard negative examples}\footnote{Hard-negatives are additional samples that either contain subtle visual changes in the image and/or subtle linguistic/semantic difference in the caption and are sampled as negatives in the same batch.}, either in the text space~\citep{kalantidis2020hardnegativemixingcontrastive, yuksekgonul2023when, zhang2024contrastingintramodalrankingcrossmodal,doveh2023teachingstructuredvisionlanguageconcepts,paiss2023teaching} -- to improve sensitivity to the order of words and subtle textual differences -- or the image space~\citep{awal2024visminvisualminimalchangeunderstanding, le2023cococounterfactuals, zhang2024countercurateenhancingphysicalsemantic} -- to improve sensitivity to subtle visual differences. 
Although these methods have somewhat improved CLIP-like models' performance on scene compositionality benchmarks~\citep{Parcalabescu_2022,zhao2022vl,yuksekgonul2023when,hsieh2023sugarcrepefixinghackablebenchmarks}, they do not explicitly address the binding problem as they focus mainly on enhancing the model’s representation capabilities with additional data, hindering their generalization to unseen scene compositions.\looseness-1

Yet, the literature on object-centric representation learning~\citep{eslami2016attendinferrepeatfast, greff2020multiobjectrepresentationlearningiterative, 
locatello2020objectcentriclearningslotattention, wu2023slotdiffusionobjectcentricgenerativemodeling, seitzer2023bridginggaprealworldobjectcentric} has long focused on devising methods to address the segregation and representation problems as a way to facilitate the subsequent compositional processing of images. This has led to the development of inductive biases to segregate different objects in a scene into distinct representational \emph{slots}, which have been shown to naturally scale to an increasing number of visual objects and relations~\citep{locatello2020objectcentriclearningslotattention, NEURIPS2023_e3cdc587, mondal2024slotabstractorsscalableabstract, savi++}. 
To the best of our knowledge, advances in object-centric representation learning are yet to be explored in the vision-language domain.\looseness-1

Therefore, in this paper, we focus on enhancing the compositional scene understanding of CLIP-like models by leveraging advances from object-centric representation learning. In particular, we propose to endow CLIP-based vision-language architectures with segregation and composition capabilities. Our core idea is to adapt the slot-centric representation paradigm for CLIP architectures and dynamically align each representational slot with the object entities mentioned in the text. To do so, we design a binding module that connects a scene graph, derived from the textual description, with a slot-structured image representation. We utilize the scene graph's relationships as constraints to effectively capture the complex interactions among the visual entities represented as slots. Our enhanced model, which we refer to as Object-Centric CLIP (OC-CLIP), not only boosts CLIP's performance in understanding multi-object compositional scenes but also improves the sample efficiency of the model when trained from scratch.\looseness-1 

Our contributions are summarized as follows:
\begin{itemize}[noitemsep]
    \item We introduce OC-CLIP, a model which endows CLIP-based architectures with segregation and composition capabilities to address the binding problem.\looseness-1
    \item We evaluate the sample efficiency of our approach against methods leveraging hard negative augmentations in a controlled 3D environment and show the overall efficiency of OC-CLIP compared to both text and image based a hard-negative augmentations.\looseness-1
    \item We demonstrate that OC-CLIP significantly enhances the binding of object-centric attributes and spatial relationships across a representative set of challenging real-world compositional image-text matching benchmarks. Notably, we report an increase of $\textbf{16.5\%}$ accuracy in the challenging \emph{swap-attribute} split of SugarCrepe compared to OpenCLIP~\citep{ilharco2021openclip} finetuned in-domain, and go from random chance to more than $\textbf{89\%}$ on COCO-spatial and $\textbf{92\%}$on GQA-spatial from the Whatsup benchmark~\citep{kamath2023whatsupvisionlanguagemodels}.\looseness-1
    \item We show the scaling potential of OC-CLIP when trained from scratch on noisy data~\citep{changpinyo2021cc12m, sharma-etal-2018-conceptual} datasets. We report an increase of \textbf{12.7\%} accuracy in zero-shot ImageNet classification compared to OpenCLIP.
\end{itemize}

\section{Related Work}
\myparagraph{Contrastive Pretraining of VLMs.}
Vision-language models (VLMs) have made substantial strides in both the vision and multi-modal domains~\citep{bordes2024introduction}. Modern VLMs are pretrained on vast, diverse and oftentimes noisy multi-modal datasets~\citep{changpinyo2021cc12m,schuhmann2022laion5bopenlargescaledataset,ilharco2021openclip, zeng2022multigrainedvisionlanguagepretraining}, and have shown substantial improvements when applied to various zero-shot tasks. CLIP~\citep{radford2021learning} presented a contrastive learning approach used for pretraining, which involves training the model to differentiate between similar and dissimilar image-text pairs. This approach encourages the model to learn a shared representation space for images and text, where semantically similar pairs are close together and dissimilar pairs are far apart. Following CLIP's lead, image-text contrastive learning has become a prevalent strategy for VLM pretraining~\citep{liu2023llava,cai2023vipllava,liu2024llavanext, dai2023instructblipgeneralpurposevisionlanguagemodels, zhai2022litzeroshottransferlockedimage, chen2022pali, beyer2024paligemmaversatile3bvlm, finiimproved}. Contrastive vision-language pretraining spans numerous downstream applications, including zero-shot image classification~\citep{zhai2022lit,radford2021learning, metzen2024autoclipautotuningzeroshotclassifiers,gao2021clipadapterbettervisionlanguagemodels},  text-to-image generation~\citep{podell2023sdxlimprovinglatentdiffusion,abdal2021clip2styleganunsupervisedextractionstylegan, ramesh2022hierarchical,saharia2022photorealistic}, as well as assessing text-image alignment~\citep{hessel-etal-2021-clipscore, cho2023finegrainedimagecaptioningclip}. In this work, we are particularly interested in the ability of CLIP-based models to evaluate compositional text-image alignment.\looseness-1

\myparagraph{Compositional Understanding Benchmarks.} 
Several benchmarks have been developed to assess the compositional understanding of VLMs. In this work, we focus on benchmarks structured as cross-modal retrieval tasks where the model needs to distinguish between correct and incorrect text descriptions given an image, and evaluations are based on accuracy metrics. The majority of these benchmarks~\citep{zhao2022vl, yuksekgonul2023visionlanguagemodelsbehavelike, Parcalabescu_2022} rely on the rule-based construction of negative captions and the generation of their associated image counter-factuals~\citep{zhang2024countercurateenhancingphysicalsemantic, awal2024visminvisualminimalchangeunderstanding}. Yet, many of these benchmarks may be solved by leveraging the language prior exclusively~\citep{vqa2, lin2024revisiting}, hence disregarding the information from the visual input. To address this, benchmarks such as SugarCrepe~\citep{hsieh2023sugarcrepe} leverage large language models to generate plausible and linguistically correct hard negatives, and show that previously introduced text-based hard negative strategies are not always effective~\citep{yuksekgonul2023when} -- \textit{e.g.}, when considering attribute and object swaps between textual descriptions.
Other benchmarks focus on assessing the VLMs' spatial understanding~\citep{kamath2023whatsupvisionlanguagemodels, yuksekgonul2023when, zhang2024countercurateenhancingphysicalsemantic}, and propose to finetune CLIP-based models on data containing a high proportion of spatial relationships since these relationships tend to be under-represented in commonly used pretraining datasets. Interestingly,~\citet{kamath2023whatsupvisionlanguagemodels} show that even when finetuning with in-domain data containing an over-representation of spatial relationships, state-of-the-art models still exhibit a close to random chance performance. In this work, we test the hypothesis that spatial relationship failures are due to the lack of composition in the similarity score computation used to train CLIP-like models.\looseness-1

\myparagraph{Object-centric Binding Inductive Biases.}
CLIP has been shown~\citep{yuksekgonul2023visionlanguagemodelsbehavelike} to be pushed to learn disentangled, bag-of-words-style representations from the contrastive loss and the easily distinguishable negatives typically used for pretraining. Although the learned representations might be effective for objects presented in isolation, they struggle with scenes containing multiple objects~\citep{tang2023when}. For example, consider a simple scene with a green apple and a yellow banana. In this case, the model must maintain and correctly link the attributes (``green'', ``yellow'') to the objects (``apple'', ``banana''), without mixing the concepts  -- \textit{e.g.}, ``yellow apple'' or `green banana''. This exemplifies the importance of devising robust mechanisms within the CLIP architecture and/or training to accurately handle multiple objects, while preventing feature interferences. 
In this work, we focus on equipping CLIP with object-centric binding inductive biases and take inspiration from the architectures proposed in the unsupervised object-centric visual representation learning literature~\citep{
locatello2020objectcentriclearningslotattention,wu2023slotdiffusionobjectcentricgenerativemodeling,seitzer2023bridginggaprealworldobjectcentric,pmlr-v177-assouel22a}. Many recent image-only approaches follow a simple inductive bias introduced by slot attention~\citep{locatello2020objectcentriclearningslotattention}, where  an image -- encoded as a set of input tokens -- is soft partitioned into
K slots. In particular, attention maps are computed via an \textbf{inverted cross attention} mechanism~\citep{WuInvertedAttentionTC}, where the softmax is applied along the query dimension in order to induce a competition between the slots to explain different groups of input tokens. In this work, we extend these inductive biases to define text-conditioned visual slots from the input image.\looseness-1

\section{Method}
\begin{figure*}
    \centering
    \includegraphics[ width=1.0\linewidth]{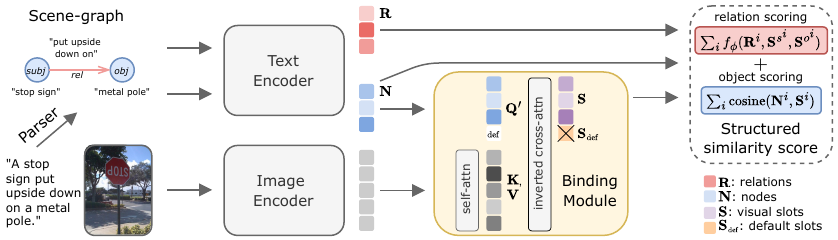}
    \caption{\textbf{Object-Centric CLIP (OC-CLIP) overview.} OC-CLIP begins with scene parsing, where we utilize a text parser (\textit{e.g.}, Llama3-based) to extract objects and relations from the input caption. The extracted text objects and relations are then fed into a text encoder, which generates distinct text embeddings for both nodes and relations. In parallel, the corresponding image is processed by an image encoder to produce patch-level image embeddings. These image embeddings are then combined with the text entity embeddings and passed through a \emph{binding module}, which outputs visual token slots embeddings. Both modality are aligned in a \emph{new space} using a structured similarity score that matches nodes embeddings to visual slots and models relational constraints between them.\looseness-1} 
    \label{fig:model}
    \vspace{-1em}
\end{figure*}

Our goal is to enhance CLIP-based architectures with object-centric binding and composition capabilities. 
Our method starts by extracting representations of distinct open-ended objects and relationships in a textual description, as well as representations of patches in an image. Next, a binding module matches the text representation of objects to the relevant image patches, producing a slot-centric representation of the image. Finally, a structured similarity score compares the slot-centric representation with the textual representations of different objects, and leverages the extracted relationships as constraints applied to the visual slots. 
Our key contributions lie in the design of the \emph{binding module}\footnote{Code  for the binding module is given in the Appendix Fig~\ref{fig:code}.} and the proposal of the \emph{structured similarity score}, which we detail in sections~\ref{ssec:bindingmod} and~\ref{sec:our_score}, respectively. Figure~\ref{fig:model} presents an overview of the proposed approach. Our approach relies on a scene-graph representation of the text modality. We assume the parser is given and 
orthogonal to our approach and discuss the choice of the parsing method in Appendix~\ref{app:parsing}.\looseness-1

\myparagraph{Notation.}
We denote as $\mathbf{x}$ an image of shape $\mathbb{R}^{h \times w \times 3}$ and as $\mathbf{\bar{x}} = [\mathbf{\bar{x}}^1, ..., \mathbf{\bar{x}}^N] = E_\phi(\mathbf{x}) \in \mathbb{R}^{N \times d} $ its patch-level encoding, where $E_\phi$ is an image encoder -- typically a pre-trained ViT~\citep{dosovitskiy2020image} -- $N$ is the number of patches and $d$ the dimensionality of the patch embeddings. We denote as $t$ the text description, or caption, associated with $\mathbf{x}$. We extract a scene graph. 
For example, the scene graph of ``A red apple to the left of a blue car'' will be represented with the set of nodes \{``red apple'', ``blue car''\} and the set of edges \{(``to the left of'', ``red apple'', ``blue car'')\}. In practice, we represent $\mathcal{N}$ as a matrix of node features $\mathbf{N}$, where each row contains the embedding of a node in the graph. Moreover, we represent each $s^i$ and $o^i$ in the relationship tuples as indices referencing the nodes (rows) in $\mathbf{N}$.\looseness-1

\subsection{Binding Module}
\label{ssec:bindingmod}
Our first contribution resides in the binding module. The idea is that when comparing the content of a caption and an image we do not want the features of different objects to interfere with each other but rather keep them separate at a representational level. The role of the binding module is thus to extract a slot-centric representation of an image where the content of the slots are pushed to represent the nodes of the associated scene graph.\looseness-1

To do so, we implement the binding module using a \textit{inverted} cross-attention layer~\citep{WuInvertedAttentionTC}, where the queries are the nodes from our scene graph and the keys and values are the image patches. We normalize the attention coefficients over
the queries' dimension in order to introduce a competition between queries to explain different parts of the visual input.
We follow common practice and set the attention's softmax temperature to $\sqrt{D}$, with $D$ being the dimensionality of the dot-product operation. 
Applying the softmax along the queries' dimension pushes all the candidate keys to be softly matched to at least one query. However, captions mostly describe specific parts of the image, and rarely capture all the visual information. Since we want only the relevant visual information to be captured by the queries, we add a set of default query tokens, stored in a matrix $\mathbf{Q}_{\text{default}}$, which participate in the competitive attention mechanism -- with the goal of absorbing the visual information not captured in the caption. These default query tokens are dropped in the subsequent computation steps of our model (akin to registers in ViT backbones~\citep{darcet2024visiontransformersneedregisters}). We find the default query tokens crucial to stabilize the training our model.\looseness-1

The binding module computations are formalized as follows:
\begin{align}
\mathbf{Q} &= \mathbf{W}_q \mathbf{N}, \nonumber\\
\mathbf{K},\mathbf{V} &= \mathbf{W}_k\mathbf{\bar{x}},\mathbf{W}_v \mathbf{\bar{x}}, \nonumber\\
\mathbf{Q'} &= [\mathbf{Q}; \mathbf{Q}_{\text{default}}], \nonumber\\
\text{Attn}(\mathbf{Q'}, \mathbf{K}, \mathbf{V}) &= \text{softmax}\left(\frac{\mathbf{Q'} \cdot \mathbf{K}^T}{\sqrt{D}}, \text{dim='Q'}\right) \cdot \mathbf{V},\nonumber\\
\mathbf{S}, \mathbf{S}_{\text{default}}&= \text{Attn}(\mathbf{Q}', \mathbf{K}, \mathbf{V}).
\end{align}
Here, $\mathbf{W}_q$, $\mathbf{W}_k$, and $\mathbf{W}_v$ are the linear projection weight matrices for the queries, keys, and values, respectively, $\mathbf{S}$ are the visual slots, $\mathbf{S}_\text{default}$ are the visual slots from default query tokens, which are discarded for subsequent steps, and [.] denotes the concatenation operation.

Thus, the output of this binding module are the visual slots $\mathbf{S}$. Intuitively, these slots are pushed to represent the visual objects, or entities, that correspond to the nodes of the scene graph. Their object-centric learning is driven by the structured similarity that we detail in the next section.\looseness-1

\subsection{Structured similarity score}\label{sec:our_score}
Our second contribution resides in the introduction of a structured similarity score, whose goal is to promote the constraints imposed by the scene graph on the learnable visual slots. Our proposed structured similarity score is composed of an \emph{object scoring} function and a \emph{relationship scoring} function. The object scoring function assesses the presence of each node in the scene graph (objects present in the caption). We model this function as the sum of the cosine similarity between each textual node representation $\textbf{N}^i$ and its assigned visual slot $\textbf{S}^i$. The relationship scoring function encourages the relational constraints imposed by each edge in the scene graph and is defined as a learnable function $f_\phi$ of the relationship embedding $\mathbf{r}^{i}$, and the visual slot representations $\textbf{S}^{s^i}$ and  $\textbf{S}^{o^i}$ corresponding to the subject and object of the relationship, respectively.
We derive the overall structured similarity score over the visual slots $\mathbf{S}$ from an image $\mathbf{x}$ and a graph $\mathcal{G} = (\{N^i\}_{i=1..M}, \{(\mathbf{r}^i, s^i, o^i)\}_{i=1..P})$ such that: $S(\mathbf{x}, \mathcal{G}) = \frac{ \alpha \sum_{i=1..M}\text{cosine}(\mathbf{N}^i, \mathbf{S}^i) + \beta\sum_{i=1..P}f_{\phi}(\mathbf{r}^i, \mathbf{S}^{s^i}, \mathbf{S}^{o^i})} {\alpha M + \beta P} $\label{eq:score}
where $\alpha$ and $\beta$ are \new{learned} parameters controlling the strength of each score. $M$ and $P$ are the number of nodes and relationships in the scene graph $\mathcal{G}$, respectively.

We define $f_{\phi}$ as follows:
\begin{equation}\label{add_mlp}
  f_\phi(\mathbf{r}, \mathbf{S}^s, \mathbf{S}^o) = \text{cosine}\left(\mathbf{r}, f_s([\mathbf{r}, \mathbf{S}^s]) + f_o([\mathbf{r}, \mathbf{S}^o])\right),
\end{equation}
where [.] denotes the concatenation of two vectors and $f_s$ and $f_o$ are MLPs that reduce the dimensionality of their inputs. 
Note that we model the relationship scoring function so that it keeps the same scale as the object scoring function and can take the order of the relationship into account. 

\subsection{Training}
The model is trained using the following loss:
\begin{equation}
 \mathcal{L} =  \mathcal{L}_{itc} + \mathcal{L}_{rel}.
\end{equation}
$\mathcal{L}_{itc}$ is the image-text contrastive loss defined to minimize the distance between image and scene graph representations from paired text-image data while maximizing the distance between image and scene graph representations from unpaired text-image data as: 
\ifarxiv
\begin{equation}
 \mathcal{L}_{itc} = - \sum_{i=1}^B \left ( \log\frac{\exp^{S(\mathbf{x}_i, \mathcal{G}_i)}}{\sum_{j=1}^B \exp^{S(\mathbf{x}_j, \mathcal{G}_i)} }  + \log\frac{\exp^{S(\mathbf{x}_i, \mathcal{G}_i)}}{\sum_{j=1}^B \exp^{S(\mathbf{x}_i, \mathcal{G}_j)} } \right ),
\end{equation}
\fi
where $B$ is the number of elements in the batch. Note that the $S$ is the structured similarity score defined in Eq.~\ref{eq:score}. \looseness-1
$\mathcal{L}_{rel}$ is the loss that pushes the model to learn a non-symmetric relationship scores: 
\begin{equation}
  \mathcal{L}_{rel} = - \sum_{i=1}^B  \log\frac{\exp^{S(\mathbf{x}_i, \mathcal{G}_i)}}{\exp^{S(\mathbf{x}_i, \mathcal{G}_i)} + \exp^{S(\mathbf{x}_i, \bar{\mathcal{G}_i})} +\exp^{S(\mathbf{x}_i, \tilde{\mathcal{G}_i)}}},
\end{equation}
where $\bar{\mathcal{G}}$ and $\tilde{\mathcal{G}}$ are altered scene graphs. In $\bar{\mathcal{G}}$, we swap the order of the subject and the object of a relationship, whereas in $\tilde{\mathcal{G}}$, we randomly chose the relationship's subject and object from the nodes in the scene graph. \new{We ablate the main components of OC-CLIP in Table \ref{tab:ablation-main}  and give a more extensive ablation analysis in Appendix \ref{sec:abla}}\looseness-1

\section{Results}
We evaluate OC-CLIP's inductive biases in 3 different settings:

\begin{itemize}[leftmargin=*,noitemsep]
    \item \textbf{Addressing CLIP's binding problem.} We show the efficiency of OC-CLIP in addressing the binding problem compared to hard-negative based augmentation on a synthetic dataset.(Section~\ref{sec:PUG}). 
    \item \textbf{Compositional understanding.} We showcase OC-CLIP's compositional understanding on real-world object-centric attribute binding and spatial relationship understanding benchmarks (Section~\ref{sec:comp}).\looseness-1 
    \item \textbf{Scaling on noisy data.} We show that OC-CLIP consistently outperforms a CLIP-based model in both zero-shot single object classification and zero-shot compositional understanding multi-object text retrieval, when training both models \emph{fully} from scratch on larger-scale and noisy dataset (Section~\ref{sec:noisy}).\looseness-1
\end{itemize}


\subsection{Addressing CLIP's bag-of-words behavior}
\label{sec:PUG}

In this section, we aim to assess the efficiency and effectiveness of leveraging hard-negatives in OC-CLIP and CLIP-like models in addressing the binding problem. To do so, we use a synthetic dataset with a closed-set vocabulary, from which we can enumerate all possible \emph{object-attribute} conjunctions and systematically evaluate the potential of CLIP-like models and OC-CLIP in addressing \emph{simple} swap-attribute retrieval tasks under varying hard-negative sample sizes.

\begin{figure*}[ht]
\centering
    \begin{subfigure}[b]{0.33\textwidth}
        \centering    \includegraphics[width=0.9\textwidth, height=0.15\textheight, keepaspectratio]{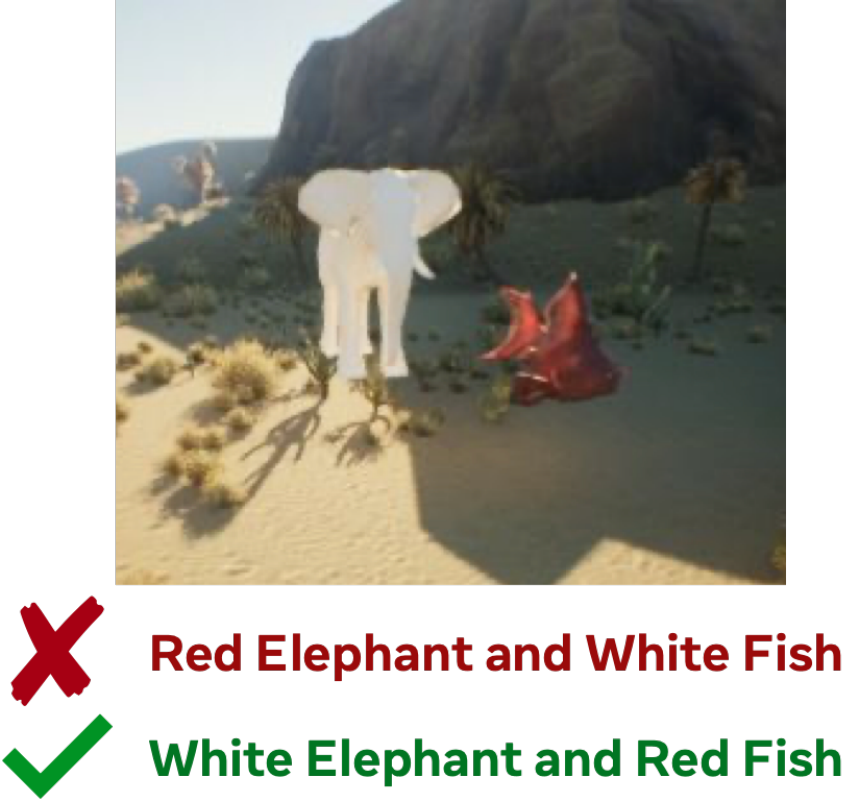}
        \caption{Synthetic data example}
        \label{fig:binding_pug:image}
    \end{subfigure}%
    \hfill
    \begin{subfigure}[b]{0.33\textwidth}
        \centering    \includegraphics[width=0.9\textwidth, height=0.4\textheight, keepaspectratio]{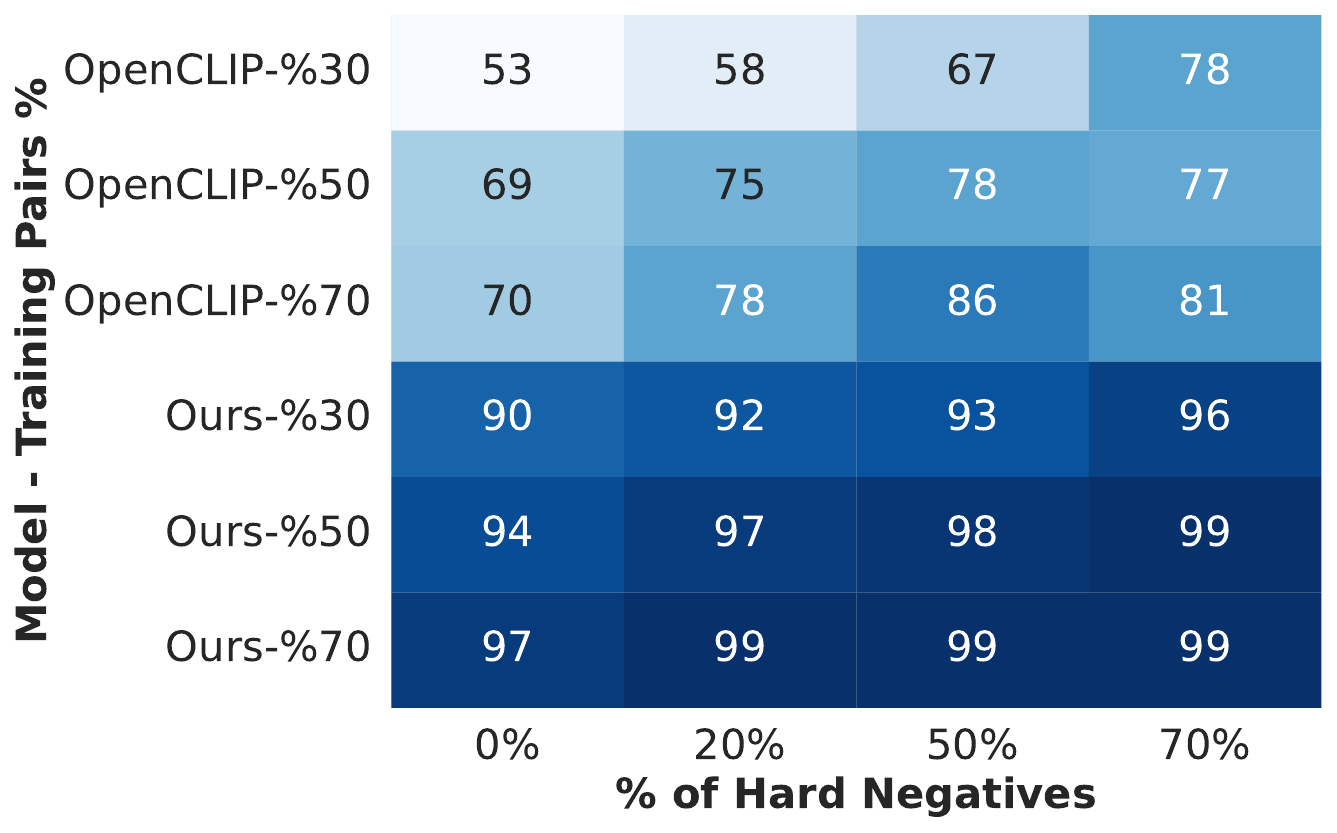}
        \caption{Seen object pairs}
        \label{fig:binding_pug:seen}
    \end{subfigure}%
    \hfill
    \begin{subfigure}[b]{0.33\textwidth}
        \centering    \includegraphics[width=0.9\textwidth, height=0.4\textheight, keepaspectratio]{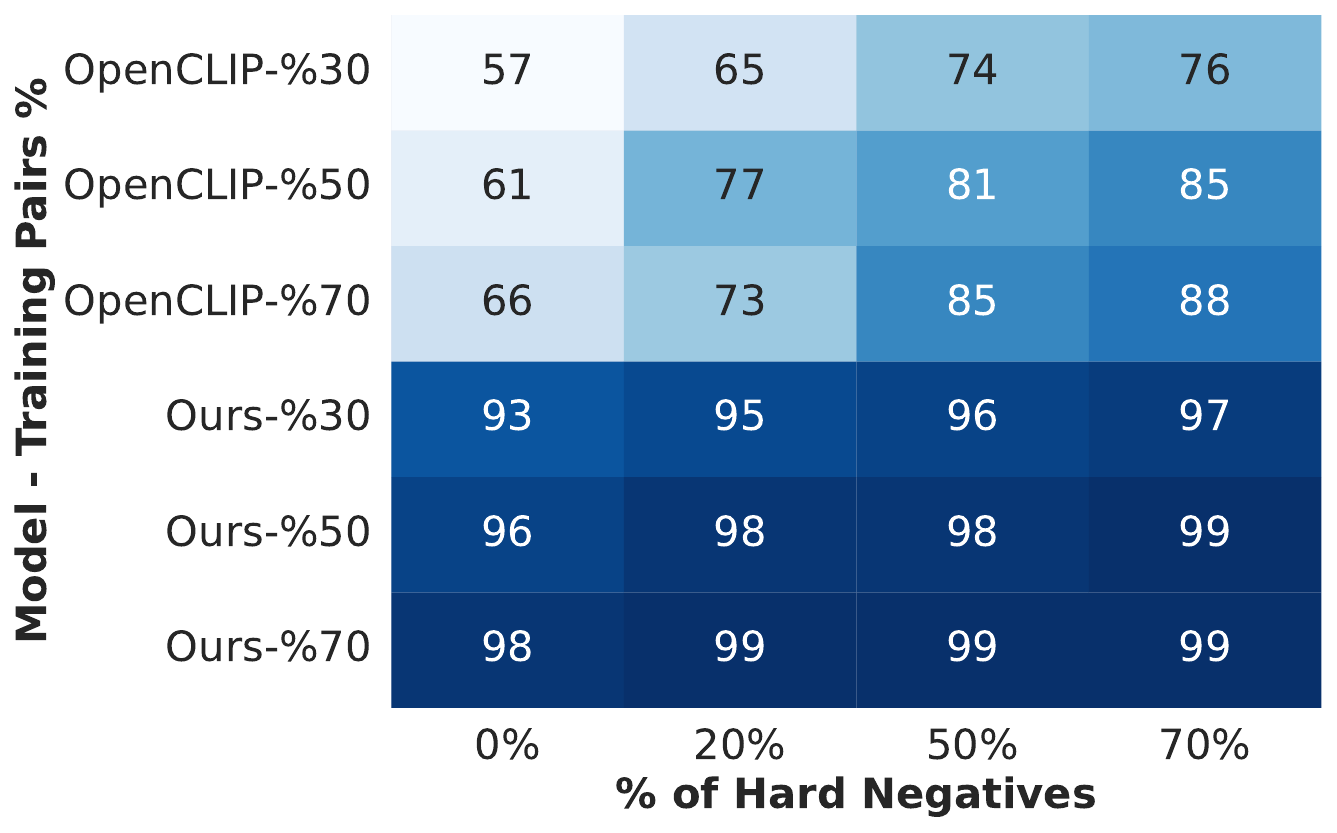}
        \caption{Unseen object pairs}
        \label{fig:binding_pug:unseen}
    \end{subfigure}
  \caption{\textbf{Efficiency and effectiveness of OC-CLIP: Analysis on synthetic data.}
  Performance of the finetuned OpenCLIP and OC-CLIP models on a binary classification task between a caption and its corresponding hard-negative given a synthetic image, as shown in (a). 
  Performance is shown as a function of the percentage of animal pairs (y-axis) seen during training and the proportion of hard-negatives used in the training data (x-axis). Results shown for (a) seen and (b) unseen object pairs.
  \looseness-1}
  \label{fig:binding_pug}
\vspace{-1em}
\end{figure*}

\textbf{Dataset.} We consider a controlled 3D environment based on PUG~\citep{bordes2023pugphotorealisticsemanticallycontrollable} and build a dataset composed of a single textured animal, or pairs of animals, in different backgrounds. We use a combination of 4 textures, 20 animal classes, and 5 different backgrounds -- \textit{e.g.}, see example in Figure~\ref{fig:binding_pug:image}. We follow prior benchmarks~\citep{hsieh2023sugarcrepe} and perform a text-retrieval task between the correct caption and the associated negative caption. We give additional details about the subsets compositions in Appendix~\ref{sec:pug_app}.\looseness-1 

\textbf{Baseline and OC-CLIP training.} We finetune models on data splits from our synthetic data, while considering an increasing proportion of hard-negative samples. We consider a CLIP model initialized with OpenCLIP weights~\cite{ilharco2021openclip}. We also initialize OC-CLIP's text and vision backbones with OpenCLIP weights, but train OC-CLIP's binding module from scratch.

\textbf{Results.} Our results, presented in Figures~\ref{fig:binding_pug:seen} and \ref{fig:binding_pug:unseen}, show that simply adding more hard-negatives to OpenCLIP's training plateaus and is not sample-efficient, as the swap-attribute binding performance always underperforms OC-CLIP trained on less data without any hard-negatives in a simple object-attribute binding task. On seen object pairs, with 70\% of the possible pairs and 70\% of their corresponding swap-attribute hard-negatives CLIP plateaus at 81\% compared to OC-CLIP which solves the task at 97\% on the same training data size and \emph{no} hard-negatives. We hypothesize that the root cause of this issue lies in the representation format used in CLIP's original formulation, which relies on a single vector to capture complex semantic relationships. Our proposed method introduces inductive biases that allow the model to learn more structured representations, avoiding superposition of features~\citep{greff2020bindingproblemartificialneural} and effectively mitigating the bag-of-words behavior. 

\subsection{Compositional Understanding}\label{sec:comp}
\label{ssec:comp_real}
In this section, we verify that the observations made in the controlled environment presented in Section~\ref{sec:PUG} also transfer to real-word datasets, thereby assessing the real-world compositional understanding of OC-CLIP.\looseness-1

\myparagraph{Datasets.}
We train OC-CLIP and finetune OpenCLIP \emph{in-domain} on a set of datasets relevant for real-world compositional understanding. 
The training text descriptions representing positive samples are taken from COCO~\citep{coco}, Visual-Genome (VG)~\citep{krishna2017visual} and GQA~\citep{hudson2019gqa}. The latter annotates images coming from Visual Genome~\citep{krishna2017visual} with objects and both spatial and non-spatial relationships, and thus contains a high representation of spatial prepositions. We evaluate the different models on the most challenging benchmarks representative of compositional understanding, ensuring that we validate both their \emph{attribute binding} and \emph{spatial relationship} understanding capabilities. In particular, we use SugarCrepe~\citep{hsieh2023sugarcrepefixinghackablebenchmarks} and ARO-A ~\citep{yuksekgonul2023visionlanguagemodelsbehavelike} for attribute binding and ARO-Relation (ARO-R)~\citep{yuksekgonul2023visionlanguagemodelsbehavelike}, COCO-spatial and GQA-spatial ~\citep{kamath2023whatsupvisionlanguagemodels} for spatial relationship understanding. Although \citet{hsieh2023sugarcrepefixinghackablebenchmarks} showed that other benchmarks such as VL-Checklist~\citep{zhao2023vlchecklistevaluatingpretrainedvisionlanguage}, COCO-Order and Flickr-Order splits of ARO~\citep{yuksekgonul2023visionlanguagemodelsbehavelike} were easily hackable because the negatives are not semantically correct, we include the results on those benchmarks for reference in Appendix~\ref{app:vl}.\looseness-1

\myparagraph{Training.}
As in section~\ref{sec:PUG}, we initialize the text and vision backbones of OC-CLIP with pre-trained model weights, and train the binding module of OC-CLIP from scratch. In particular, we initialize the text backbone with OpenCLIP 
weights~\citep{ilharco2021openclip} and consider two different vision backbones, OpenCLIP (ViT-B-16)~\citep{ilharco2021openclip} and DinoV2 (ViT-B-14)~\citep{oquab2024dinov2learningrobustvisual}, to show the flexibility of our binding module and learned structured similarity score. We noticed that taking the patches from earlier layers in OpenCLIP helps the training and ablate it in Appendix~\ref{sec:abla}. We use a batch size of 128 and a learning rate of $2\cdot 10^{-4}$ to train OC-CLIP for 100 epochs. We use a batch size of 256 -- following previous finetuning approaches~\citep{kamath2023whatsupvisionlanguagemodels, yuksekgonul2023when} -- and a learning rate of $4\cdot 10^{-6}$ for 20 epochs to finetune the OpenCLIP baseline. We run all the models for 3 seeds and report the mean performance along with their standard deviation. Note that since OC-CLIP's binding module is trained from scratch, OC-CLIP's learned vision-language-aligned space does not rely on the vision-language alignment captured by the CLS token of OpenCLIP's backbone (in fact, we drop the CLS token). Therefore, \emph{the new representation space learned by OC-CLIP can only be expected to generalize within the vocabulary it has been trained on.}\looseness-1

\myparagraph{Baselines.} We report the performance of a representative set of strong baselines which we separate in two groups: the first group of baselines are models trained contrastively and finetuned in-domain (on COCO/VG) and the second group are hard-negative-based and recaptioning-based methods, further divided into small scale and large scale. For the first group, we include OpenCLIP -- referred to as OpenCLIP-FT --, BLIP~\citep{li2023blip}, and XVLM~\citep{zeng2022multigrainedvisionlanguagepretraining}. BLIP is augmented with an image-text matching loss and XVLM uses bounding boxes to assist the object-centric binding. Note that these two baselines are also equipped with a language modeling objective which may help identify unplausible captions. 
For the second group, we select 
methods that augment the dataset with rule-based text hard-negatives (NegCLIP~\citep{yuksekgonul2023when}), language-model-based hard-negatives (CE-CLIP~\cite{zhang2020contrastive} and CLIP-SVLC~\citep{doveh2023teachingstructuredvisionlanguageconcepts}), and image-\&-language-model-based hard-negatives (CLIP-CC~\citep{zhang2024countercurateenhancingphysicalsemantic}). We also include dense recaptioning baselines such as DAC~\citep{doveh2023dense} for reference.\looseness-1

\myparagraph{Attribute Binding Results.}
We evaluate the attribute binding capabilities of OC-CLIP and baselines on SugarCrepe~\citep{hsieh2023sugarcrepefixinghackablebenchmarks} and ARO-A~\citep{yuksekgonul2023when} benchmarks. 
We report the results in Table~\ref{tab:sugarcrepe}. When comparing OpenCLIP$_{\text{FT}}$ to OC-CLIP (ours -- both models), we observe notable performance boosts on ARO-A and SugarCrepe's swap-attribute, and swap-object. In particular, {OC-CLIP\tiny{B-14}} shows a performance boost of +24\% on ARO-A, while in SugarCrepe, our model achieves improvements of +16.5\% on the hard swap attribute split, +20.4\% on the swap object split, and a smaller +4.1\% on the replacement relationship split. Moreover, both OC-CLIP models perform similarly to OpenCLIP$_{\text{FT}}$ on the remaining SugarCrepe splits. This is to be expected since the remaining splits do not require precise binding to distinguish between positive and negative captions and may therefore be solved with a bag-of-words-like representation. 
We additionally compare OC-CLIP to finetuned versions of CLIP that rely on in domain hard-negatives (NegCLIP, CE-CLIP, CC-CLIP) and with dense recaptioning (DAC-LLM and DAC-SAM). In particular DAC finetunes OpenCLIP with $\sim3$M VLM-generated dense captions (along with their corresponding hard negatives) that significantly increase the vocabulary coverage compared to methods that only finetune in domain (\textit{e.g.}, on COCO). Interestingly, OC-CLIP still outperforms them on both swap-attribute and swap-object, showing improvements of +13.6\% and +8.4\% over the second best performing method, respectively. Those results confirm the behavior that we observed in Section~\ref{sec:PUG} and the inefficiency of hard-negative methods in solving the binding problem of CLIP-like models, even at the scale of DAC finetuning.\looseness-1
\begin{table*}[ht]
\begin{small}
\begin{sc}

\ifarxiv
  \scalebox{0.92}{%
    \begin{tabular}{ccccccccc}
    \cmidrule(lr){1-9}
    \multirow{2}{*}{Model} & \multicolumn{2}{c}{Swap} & \multicolumn{2}{c}{Add} & \multicolumn{3}{c}{Replace} & \multicolumn{1}{c}{ARO} \\
    \cmidrule(lr){2-3} \cmidrule(lr){4-5} \cmidrule(lr){6-8} \cmidrule(lr){9-9}
    & Object & Attribute & Object & Attribute & Object & Attribute & Relation & Attribution \\
    \cmidrule(lr){1-9}
    \multicolumn{9}{l}{
    \textit{Zero-shot }} \\
    OpenCLIP   & 68.2 & 66.2 & 82.7& 80.3 & 93.8 & 82.8 & 67.3 & 63.2 \\
    \cmidrule(lr){1-9}
    \multicolumn{9}{l}{
    \textit{In-domain ft baselines} }\\
    BLIP & 66.2 & 76.2 & - & - & \textbf{96.5} & 81.9 & 68.35 & \textbf{88.0} \\
    XVLM  & 64.9 & 73.9 & - & - & 95.2 & 87.7 & 77.4 & 73.4 \\
    OpenCLIP$_{\text{FT}}$ & 63.1 \tiny{$\pm 0.6$} & 72.4\tiny{$\pm 1.1$} & \textbf{93.4} \tiny{$\pm 0.2$} & 83.1 \tiny{$\pm 0.5$} & 95.4 & 87.0 \tiny{$\pm 0.6$} & 75.5 \tiny{$\pm 0.6$} & 60.0 \\
    \cmidrule(lr){1-9}
    \multicolumn{9}{l}{
    \textit{Hard-Negative - small scale}} \\
    NegCLIP  & 75.2 & 75.4 & 88.8 & 82.8 & 92.7 & 85.9 & 76.5 & 71 \\
    CE-CLIP  & 72.8 & 77 & 92.4 & 93.4 & 93.1 & 88.8 & 79 & 76.4 \\
    CC-CLIP & 68.6 & 73.6 & 86.7  & 90.3  & 95.9  & 87.9  & 76.2 & - \\
    CLIP-SVLC& - & - & - & - & - & - & - & 73.0 \\
    \cmidrule(lr){1-9}
    \multicolumn{9}{l}{
    \textit{Hard-Negative/Dense Captioning - large scale}} \\
    DAC-LLM & 75.1& 74.1& 89.7&\textbf{97.7} & 94.4& \textbf{89.3}&\textbf{84.4}& 73.9 \\
    DAC-SAM &71.8 & 75.3& 87.5& 95.5& 91.2&85.9 &83.9& 70.5 \\
    \cmidrule(lr){1-9}
    \multicolumn{9}{l}{\textit{Ours} }
    \\
    OC-CLIP \tiny{B-16}  & 76.6 \tiny{$\pm 0.6$} & 87.5 \tiny{$\pm 0.5$} & 91.1\tiny{$\pm 0.4$}  & 83.8 \tiny{$\pm 1.0$} & 94.6 \tiny{$\pm 0.4$} & 87.9 \tiny{$\pm 0.1$} & 76.0 \tiny{$\pm 0.4$}  & 83.2\tiny{$\pm 0.3$}\\
    OC-CLIP \tiny{B-14}  & \textbf{83.5} \tiny{$\pm 0.2$} & \textbf{88.9} \tiny{$\pm 0.6$} & 92.8\tiny{$\pm 0.1$} & 84.8 \tiny{$\pm 0.1$} & 95.9 \tiny{$\pm 0.4$} & 89.2\tiny{$\pm 0.1$}  & 79.6 \tiny{$\pm 0.3$} & 84.0\tiny{$\pm 0.$}\\
    \cmidrule(lr){1-9}
    \end{tabular}
    }
\fi

\end{sc}
\end{small}
\caption{\textbf{Attribute binding: Performance on SugarCrepe and ARO-A.} Both OpenCLIP-FT and OC-CLIP are initialized with the same OpenCLIP checkpoints. OC-CLIP is trained with two ViT base backbones with different resolutions: OpenCLIP's backbone (B-16) and Dinov2 (B-14).OC-CLIP's bidning module is always trained from scratch.\looseness-1 }
\label{tab:sugarcrepe}
\end{table*}\looseness-1

\myparagraph{Relationship Understanding Results.}
We evaluate the spatial relationship understanding capabilities of OC-CLIP and baselines on COCO-spatial, GQA-spatial, and ARO-Relation (ARO-R). Note that ARO-Relation contains both spatial and non-spatial relations but about half of the test examples consists of left/right relationships understanding. We report the results in Table~\ref{tab:rel} and show consistent improvements of both OC-CLIP models over the baseline models and across the 3 datasets. In particular, the best OC-CLIP model outperforms OpenCLIP-FT by +44.1\% on COCO-spatial, +43.6\% on GQA-spatial, and +34.8\% on ARO-R. When compared to contrastive VLMs finetuned with in-domain data (XVLM, BLIP), OC-CLIP models exhibit superior performance, with improvements between +10\% and +27\% over the strongest contrastive finetuned VLM. Finally, when compared to baselines leveraging hard-negatives (NegCLIP), OC-CLIP remains the highest performer. Additional results on the ARO benchmark are reported in Table~\ref{tab:vl_aro} of the Appendix.\looseness-1

\begin{table}
\begin{small}
\begin{sc}
  \centering
  \scalebox{0.9}{%
  \begin{tabular}{c c c c}
    \toprule
    Model & COCO-spatial & GQA-spatial& ARO-R \\
    \cmidrule(lr){1-1}\cmidrule(lr){2-4}
    XVLM & 73.6 & 67  & 73.4\\
    BLIP & 56.4 & 52.6 & 59\\
    NegCLIP & 46.4 & 46.7& 80.2 \\
    OpenCLIP$_{\text{FT}}$&45.6 \tiny{$\pm 0.2$}&49.1\tiny{$\pm 1.1$} & 50.1\tiny{$\pm0.4$}\\
    \cmidrule(lr){1-1}\cmidrule(lr){2-4}
    OC-CLIP \tiny(B-16) & 86.3 & 90.0 & 84.3 \\
    OC-CLIP \tiny(B-14) & \textbf{89.7} & \textbf{92.7} &\textbf{84.9} \\
    \bottomrule
  \end{tabular}%
  }
  \caption{\textbf{Spatial relationship understanding: Performance on COCO-spatial, GQA-spatial from the Whats'up Benchmark and ARO-R.} We finetune both OpenCLIP (OpenCLIP$_{\text{FT}}$ here) and OC-CLIP in-domain on COCO, Visual Genome, and GQA data. Both models are initialized with the same OpenCLIP checkpoints.\looseness-1 }
  \label{tab:rel}
  \end{sc}
  \end{small}
  \looseness-1
\end{table}
\subsection{Training OC-CLIP from scratch}\label{sec:noisy}
In this section, we aim to assess the potential of OC-CLIP when trained \emph{fully} from scratch from scene-graphs obtained from large scale non-human-curated captioning dataset.\looseness-1

\textbf{Datasets.} We train both ViT-B-16 OpenCLIP model and OC-CLIP \emph{fully} from scratch on increasingly large dataset sizes using CC3M~\citep{sharma-etal-2018-conceptual}, CC12M~\citep{changpinyo2021cc12m} and the combination of both datasets. We evaluate all models on ImageNet~\citep{Deng2009ImageNetAL} zero-shot classification in this section, and report results on the ELEVATER suite~\citep{li2022elevaterbenchmarktoolkitevaluating} in Appendix (Table~\ref{tab:classification-downstream}). We also evaluate zero-shot compositional understanding of the models on the challenging swap-object and swap-attribute splits of SugarCrepe, and on Winoground~\citep{thrush2022winoground}.\looseness-1

\textbf{Baseline and OC-CLIP training.} Both CLIP and OC-CLIP architectures are trained \emph{fully} from scratch for 5, 15, or 25 epochs, using a batch size of 4096, a learning rate of $1\cdot 10^{-3}$, 2k steps of learning rate warm-up, and a cosine decay. As recommended by~\citet{mu2021slipselfsupervisionmeetslanguageimage}, we use AdamW optimizer with $0.5$ of weight decay and $\beta_2$ set to 0.98. We use a ViT-B-16 backbone for both models. Since OC-CLIP's text bakcbone only needs to encode single objects and relationships, we use a smaller text bakcbone with a context length of $20$ and only $6$ layers instead of 12 . Note that we do not tune the hyper-parameters in this experiment. We further discuss those design choices in Appendix~\ref{sec:scale}. 

\sidecaptionvpos{figure}{t}
\begin{SCfigure*}[.55]
\begin{wide} 
    \centering
    \subfloat[Sample efficiency]{
        \centering
        \includegraphics[height=4.5cm]{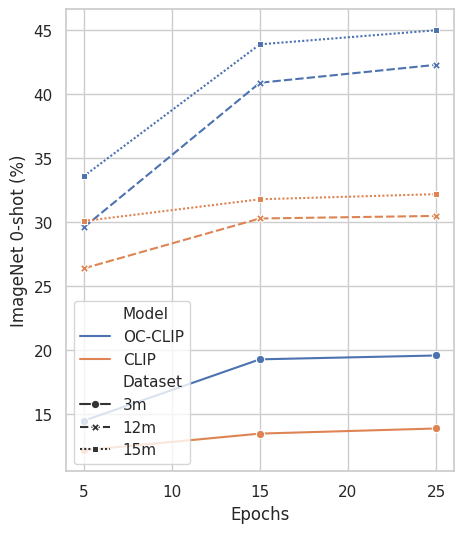}
        \label{fig:noisy_eff}}
    \hfill
    \subfloat[Zero-shot accuracy]{
        \centering
        \includegraphics[height=4.5cm]{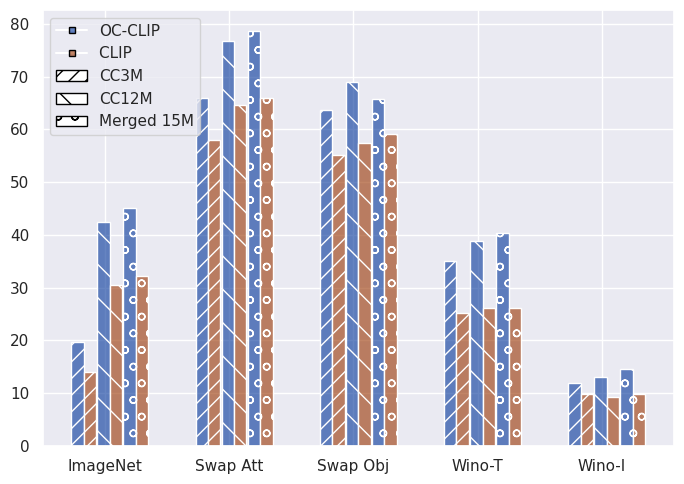}
        \label{fig:noisy_cls}}
    \hspace{-.2cm}
    \hfill
    \caption{\textbf{Scaling the training on noisy data.} CLIP and OC-CLIP  are trained \emph{from scratch} on varying sizes of data (3M, 12M and 15M) 
    for a varying number of epochs. OC-CLIP shows (b) better zero-shot compositional understanding performance on SugarCrepe's swap-attribute and swap-object, and on Winoground (I = Image score and T = Text score), as well as (a) better sample efficiency shown on zero-shot ImageNet classification.\looseness-1}
    \label{fig:noisy}
\end{wide} 
\end{SCfigure*}

\textbf{Results.} We start by verifying the sample efficiency of OC-CLIP using ImageNet~\citep{Deng2009ImageNetAL} zero-shot classification performance in Figure~\ref{fig:noisy_eff}. We show that OC-CLIP shows better sample-efficiency than the baseline trained on the same data, while using a smaller text backbone. We then evaluate OC-CLIP on zero-shot classification and compositional understanding in Figure~\ref{fig:noisy_cls}. Interestingly, OC-CLIP shows performance gains in general zero-shot classification ($+12.8\%$ on ImageNet, when trained from scratch on CC3M+CC12M) while also showcasing substantial improvements in zero-shot compositional understanding. For example, OC-CLIP exhibits a notable $+12.7\%$ and $+6.6\%$ in SugarCrepe's swap-attribute and swap-object splits, respectively. This experiment shows that the structured training of OC-CLIP is also effective when scaling to noisy image-caption dataset and, therefore, does not solely rely on high-quality human captions. 
We additionally report extensive zero-shot downstream classification performance on the ELEVATER~\citep{li2022elevaterbenchmarktoolkitevaluating} suite and discuss the computation trade-off of our approach in Appendix~\ref{sec:scale}.
We leave further scaling for future work.
\subsection{Ablations}
In Table \ref{tab:ablation-main} we ablate the key design choices of our model. Specifically, we investigate two key components of the model: the use of competitive (inverted) cross-attention and the local graph contrastive loss. On the one hand, results show that removing the competitive cross-attention mechanism greatly affects fine-grained attribute binding (decreasing from $89.0$ to $85.9$). On the other hand, removing the local graph contrastive loss significantly impacts downstream relational understanding, with accuracy decreasing from $80.5$ to $72.8$. Adding attention layers helps relational understanding (boosting performance from  $77.6$ to $80.5$), while adding more default tokens does not necessarily help with attribute binding. These findings highlight the importance of the main design choices behind OC-CLIP. More extensive ablations are presented in Appendix~\ref{sec:abla}.\looseness-1
\begin{table}[htbp]
\begin{small}
\begin{sc}
  \centering
  \scalebox{0.8}{%
  \begin{tabular}{cccc|cc}
    \toprule
     Loc Loss & Comp. X-Att & Attn Lay & Default & Rel & Att \\
    \midrule
     \checkmark & \checkmark &  \checkmark &1 & 80.5 & 89.0 \\
    \midrule
     \checkmark & \checkmark  & \checkmark& 4 & 79.2 & 87.6 \\
      -& \checkmark & \checkmark &1 &  72.8 & 87.7 \\
      \checkmark & - & \checkmark &- & 78.3 & 85.9 \\
      \checkmark & \checkmark  &-& 1 &77.6 & 87.8 \\
    \bottomrule
  \end{tabular}%
  }
  \caption{\textbf{Ablation of OC-CLIP's main components.} Fine-grained accuracy on attribute binding and relational splits  of SugarCrepe.\looseness-1}
  \label{tab:ablation-main}
  \end{sc}
  \end{small}
  \vspace{-2em}
\end{table} 

\section{Conclusion and limitations}

\myparagraph{Conclusion.} In this paper, we proposed Object-Centric CLIP (OC-CLIP), a method to enhance the compositional scene understanding of CLIP-like models by leveraging advances from object-centric representation learning. Our approach adapts the slot-centric representation paradigm to CLIP and dynamically aligns each representational slot with the objects mentioned in the text description. This is achieved by the introduction of a binding module and a structured similarity score that allows to train OC-CLIP in a contrastive way. We evaluated our approach against common hard-negative augmentation strategies and demonstrated that OC-CLIP significantly enhances the binding of object-centric attributes and spatial relationships across a representative set of challenging real-world compositional image-text matching benchmarks. Notably, we reported an increase of +16.5\% accuracy in the challenging swap-attribute split of SugarCrepe compared to OpenCLIP finetuned with in-domain data and drastically improved performance on COCO-spatial and GQA-spatial from the Whatsup benchmark, moving from random chance to more than $89\%$. Finally we show the scaling potential of OC-CLIP to be trained from scratch on a noisy dataset~\citep{changpinyo2021cc12m, sharma-etal-2018-conceptual} datastet. Notably we report performance gain in zero-shot classfication ($+12.8\%$ in ImageNet \ref{tab:classification-downstream}) while maintaining a  significant  gap in zero-shot SugarCrepe swap attribute ($+12.7\%$) and swap obj ($+6.6\%$) splits.\looseness-1

\myparagraph{Limitations and Future Work.} 
Our proposed Object-Centric CLIP (OC-CLIP) model has several limitations and avenues for future work. Notably, our approach relies on a parser to extract object-centric attributes and spatial relationships from text descriptions. While we have chosen an LLM-based parser, which is discussed in Appendix~\ref{app:parsing}, studying the different biases of LLM-based parser families could be interesting. A related promising direction for future research is also to explore the possibility of parsing scene graphs directly from Visual Language Models (VLMs), using both visual and textual inputs. Additionally, we plan to investigate the synergy between long-captioning and our scene graph-based training approach, aiming to study the complementary strengths of these two data-centric and model-centric paradigms.

\bibliographystyle{plainnat}
\bibliography{iclr2025_conference}
\clearpage
\onecolumn

\appendix
\section{Appendix}
\subsection{Ablations}\label{sec:abla}
In this section we ablate and discuss some important design choice of OC-CLIP. We separately ablate and discuss : 
\begin{itemize}
    \item The \textbf{similarity score } coefficients $\alpha$ and $\beta$ that control the weight of the objects and relations in the global graph-image similarity score.
    \item \textbf{Binding module inductive biases} and their impact on compositional understanding performance. 
    \item \textbf{Local Loss} impact on downstream compositional
    understanding of relationships.
    \item \textbf{Layer selection} with OpenCLIP backbone.
\end{itemize}

\paragraph{Similarity Score}
OC-CLIP's structured global similarity score is a combination of the object and relationship components respectively weighted by two learnt parameters $\alpha$ and $\beta$ balancing the different contributions. We let the model learn those parameters throughout the training. However, during preliminary experiments we tested a different combinations of initial coefficient within the [1.5, 1, 0.5, 0.1] grid and noticed that the model was always converging to a $\frac{\alpha}{\beta} \sim 3$ without any difference in the downstream compositional performance. We thus fix the initial coefficients to $\alpha=1.5$ and $\beta=0.5$ and treat them as parameters.

\paragraph{Default Token and Competitive Cross Attention}
In the binding module we propose to use an inductive biases to encourage the query tokens to attend to different groups of patches. In order to do so we use a competitive attention mechanism, the so called inverted cross attention common to many object-centric image encoder architecture \citep{locatello2020objectcentriclearningslotattention, WuInvertedAttentionTC}. We found that the use of inverted cross attention impacts slightly the fine-grained attribute binning performance (see ATT performance in Table~\ref{tab:ablation-main}), -Comp Att model does not use any inverted cross attention and is rather implemented with a regular cross attention mechanism, the softmax being done along the keys dimensions.). The finegrained attribute understanding (ATT) is affected by the absence of competitive attention between query slots going from 89.0\% to 85.9\% accuracy.

\paragraph{Local Graph Contrastive Loss} In designing the structured similarity score of OC-CLIP the relational component is formulated as the following cosine similarity $f_\phi(\mathbf{r}, \mathbf{S}^s, \mathbf{S}^o) = \text{cosine}(\mathbf{r}, f_s([\mathbf{r}, \mathbf{S}^s]) + f_o([\mathbf{r}, \mathbf{S}^o])$. In theory both $f_s([\mathbf{r}, \mathbf{S}^s]) $ and  $f_o([\mathbf{r}, \mathbf{S}^o]) $  can collapse to ignore the subject  object visual representation. In order to prevent such collapse we propose to add a local graph contrastive loss that shares similarity with hard-negative based learning. We enforce the model to model with a higher similarity the graph composed of the same nodes but with either swapped object and subject indices or shuffle objects and subjects indices within the local graph. In both of those cases the relation component of the structured similarity score becomes (for a single relation graph) : 
\begin{equation}
    \text{swap } \tilde{G}; \text{cosine}(\mathbf{r}, f_s([\mathbf{r}, \mathbf{S}^s]) + f_o([\mathbf{r}, \mathbf{S}^o])\\
\end{equation}
\begin{equation}
    \text{swap } \tilde{G}; \text{cosine}(\mathbf{r}, f_s([\mathbf{r}, \mathbf{S}^o]) + f_o([\mathbf{r}, \mathbf{S}^s])\\
\end{equation}
\begin{equation}
     \text{shuffle } \bar{G} ; \text{cosine}(\mathbf{r}, f_s([\mathbf{r}, \mathbf{S}^{j!=s}]) + f_o([\mathbf{r}, \mathbf{S}^{i!=o}])
\end{equation}
This prevents the model from collapsing because  ground-truth $G$ is distinguishable from $\tilde{G}$ and $\bar{G}$ only if the visual representations are not ignored in the relationships components.
As shown in Table~\ref{tab:ablation-main}, removing the local loss effectively impacts downstream relational understanding on SugarCrepe with a REL accuracy decreasing from $80.5$ to $72.8$ hence showing the effectiveness of the local graph contrastive loss.

\begin{figure}
    \centering
    \includegraphics[width=\textwidth]{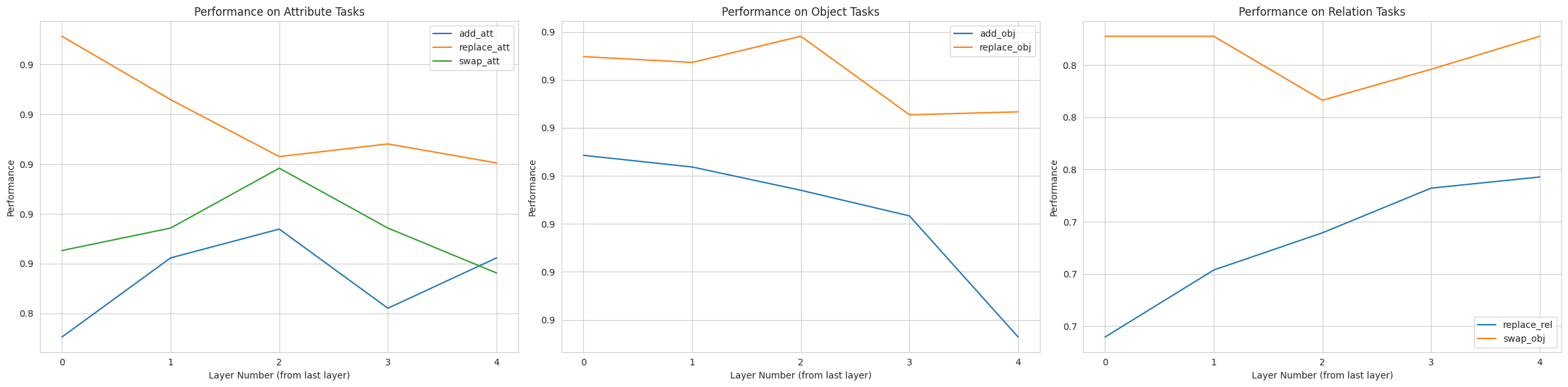}
    \caption{ViT features layer ablation.}
    \label{fig:layers}
\end{figure}

\begin{figure}[ht]
\centering
    \begin{subfigure}[b]{0.19\textwidth}
        \centering    \includegraphics[height=1in]{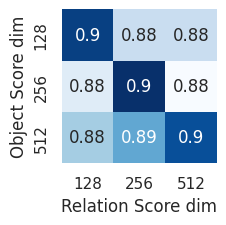}
        \caption{Swap Att}
    \end{subfigure}%
    \hfill
    \begin{subfigure}[b]{0.19\textwidth}
        \centering    \includegraphics[height=1in]{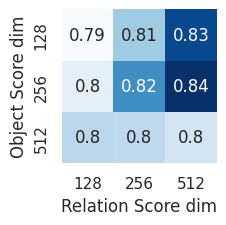}
        \caption{Swap Obj}
    \end{subfigure}%
    \hfill
    \begin{subfigure}[b]{0.19\textwidth}
        \centering    \includegraphics[height=1in]{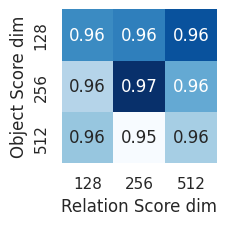}
        \caption{Replace Obj}
    \end{subfigure}%
    \hfill
    \begin{subfigure}[b]{0.19\textwidth}
        \centering    \includegraphics[height=1in]{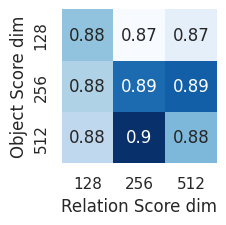}
        \caption{Replace Att}
    \end{subfigure}%
    \hfill
    \begin{subfigure}[b]{0.19\textwidth}
        \centering    \includegraphics[height=1in]{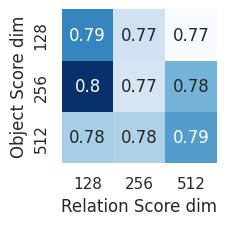}
        \caption{Replace Rel}
    \end{subfigure}%
  \caption{\textbf{Score dimensionality ablations} In this ablations we keep the initialization seed fixed and vary the dimensionality of the relation score $d_{\text{rel}}$ (x-axis) and object score $d_{\text{obj}}$(y-axis) and report the performance on the swap and replace splits of sugarcrepe.}
  \label{fig:abla_dim}
\end{figure}

\paragraph{Scoring dimensionality} Our structured similarity score allows the text encoder to focus on encoding information about individual objects and their relationships, rather than the entire scene configuration. To achieve this, we experimented with different dimensionality for both the object scoring bottleneck and the relationship scoring bottleneck. Specifically, each of these scores is designed as a cosine distance between a text representation and a visual component (as described in Section \ref{sec:our_score}), with each operating at a bottleneck dimension of $d_{\text{obj}}$ and $d_{\text{rel}}$. In contrast, OpenCLIP represents both the scene caption and the visual representation at a shared dimension of $d=512$. We expect that our model can operate effectively at a much lower dimensionality, as it requires less capacity to encode single objects and relationships. We present an ablation study of these two dimensions in Figure \ref{fig:abla_dim} and notice that our model is quite robust when we operate on lower dimensional space (eg. 128). We use this results to scale our experiments and train the model from scratch with a smaller text encoder as explained in the next section~\ref{sec:scale} corresponding to experiments shown in Section \ref{sec:noisy}.
\paragraph{Layer Selection} Previous work focusing on dense segmentation tasks\citep{xu2022groupvitsemanticsegmentationemerges} show that taking features from earlier layer in CLIP's ViT help with fine-grained tasks. Here we ablate OC-CLIP's version using the OpenCLIP pretrained backbone when inserting the binding module after layer $L-k$ for $k \in [0..4]$, 0 being the last layer. Results on merged Sugarcrepe splits are given in Figure~\ref{fig:layers}. The object related queries seem to decrease as a function of $k$ where as the replace rel split is increasing with k. To that end, we actually insert our binding module after layer $L-2$, where $L$ is the index of the last transformer layer of the ViT.

\subsection{Experiments on CC3M/12M.}\label{sec:scale}
\new{In the compositional understanding experiments we compare our approach with data-centric finetuning methods that do not add any additional parameters. These methods are expected to retain some of the general capabilities of the initial backbone. In contrast, our binding  and relationship modules is trained from scratch, which means it may not generalize as well to unseen data and can only be expected to work well within the vocabulary domain it has been exposed to (eg. COCO/VG/GQA in our experiments setting). 
However an interesting question would be to asses whether such inductive biases and structured similarity object might have some sclaing potential on noisy and non human curated datasets such as CC12M \citep{changpinyo2021cc12m}. To answer that question we propose to train both CLIP and OC-CLIP architectures from scratch on combinations of CC3M,  CC12M and CC3M+12M and compare both of their general understanding and compositional downstream performance.
In addition to the zero-shot evaluation, we also provide a computational analysis of the binding module to gain insights into its behavior and limitations.

\paragraph{Training Details}

 In order to show the potential of OC-CLIP to learn from scene-graph obtained from a non human-curated captioning dataset we train both ViT-B-16 OpenCLIP model and OC-CLIP from scratch on CC3M~\citep{sharma-etal-2018-conceptual}, CC12M~\citep{changpinyo2021cc12m} and the merge of both (15M) . We did not tune the hyperparameters and used the same hyperparameters as suggested in \citep{mu2021slipselfsupervisionmeetslanguageimage}. Both models are trained for 5, 15, and 25 epochs, using a batch size of 4096, a learning rate of $1e-3$, 2k steps learning rate warmup and a cosine decay after. As recommended by \citet{mu2021slipselfsupervisionmeetslanguageimage} we used AdamW optimizer with 0.5 of weight decay and $\beta_2$ set to 0.98.We report extensive zero-shot downstreeam classification performance on the ELEVATER \citep{li2022elevaterbenchmarktoolkitevaluating} suite in Table \ref{tab:classification-downstream}. We did not use any templates and use raw labels instead. Both models were trained using 4x8 V100 GPUS with a local batch size of 128. OC-CLIP shows  performance gains in both zero-shot classification (a notable +12.7\% in ImageNet) when trained on the same setting.  These experiments show that the structured training of OC-CLIP can scale to automatic alt-text captioning dataset. We leave further scaling for future work as the main focus of our work is to emphasize the binding problem that arises when using a vector-based representation and a set of inductive biases  as a way of operating on a more structured representation (eg. scene graph).
\begin{table*}
\setlength{\tabcolsep}{2pt}
\linespread{1}
\scriptsize
\begin{adjustbox}{width=\textwidth}
\new{    
\begin{tabular}{ll ccccccccccccccccccccccc|c} &&
\rotatebox[origin=lb]{90}{\smash{Food-101}} & \rotatebox[origin=lb]{90}{\smash{CIFAR-10}} & \rotatebox[origin=lb]{90}{\smash{CIFAR-100}} & \rotatebox[origin=lb]{90}{\smash{SUN397}} & \rotatebox[origin=lb]{90}{\smash{Cars}} & \rotatebox[origin=lb]{90}{\smash{Aircraft}} & \rotatebox[origin=lb]{90}{\smash{DTD}} & \rotatebox[origin=lb]{90}{\smash{Pets}} &  \rotatebox[origin=lb]{90}{\smash{FER-2013}} & \rotatebox[origin=lb]{90}{\smash{STL-10}} & \rotatebox[origin=lb]{90}{\smash{EuroSAT}} &
\rotatebox[origin=lb]{90}{\smash{RESISC45}} & \rotatebox[origin=lb]{90}{\smash{GTSRB}} & \rotatebox[origin=lb]{90}{\smash{KITTI}} & \rotatebox[origin=lb]{90}{\smash{Country211}} & \rotatebox[origin=lb]{90}{\smash{PCAM}} &
\rotatebox[origin=lb]{90}{\smash{UCF101 Frames}} & \rotatebox[origin=lb]{90}{\smash{CLEVR}} & \rotatebox[origin=lb]{90}{\smash{HatefulMemes}} & \rotatebox[origin=lb]{90}{\smash{MNIST}} & \rotatebox[origin=lb]{90}{\smash{SST2}} &
\rotatebox[origin=lb]{90}{\smash{ImageNet}} \\
\hline
\multirow{2}{1em} & CLIP (3m) & 12.8 & 44.9 & 19.9& 27.9 & 1.2 & 1.3 & 9.7 & 12.6 &1.3 &76.1 & 15.8 & 13.5 & 6.9 & 24.9 & 0.6 & 44.3 & 17.8 & 11.1 & 51.2 &7.8& 47.4 & 13.9  \\
& OC-CLIP (3m)  & 15.3 & 57.1 & 24.8 & 33.1 & 1.2 & 1.6 & 13.3& 16.3 & 9.9&82.0 & 16.2 & 22.0 & 4.5 & 30.8 & 0.7 & 55.6 & 22.5 & 12.5 & 52.0 &11.6& 50.1 & \textbf{19.2}  \\
\hline
\multirow{2}{1em} & CLIP (12m) & 42.7 & 63.2 & 30.2 & 42.7 & 15.5 & 3.1 & 14.5 & 52.6 &13.1 &85.7 & 12.3 & 28.5 & 8.3 & 34.9 & 3.9 & 54.4 & 33.9 & 11.8 & 51.4 &11.2& 51.9 & 30.5  \\
& OC-CLIP (12m)  & 54.1 & 74.5 & 44.6 & 51.4 & 21.1 & 3.7 & 19.4 & 66.4 & 7.4&91.9 & 31.8 & 40.6 & 8.1 & 41.9 & 5.9 & 56.1 & 45.7 & 12.7 & 49.2 &9.7& 50.2 & \textbf{42.3}  \\
\hline
\multirow{2}{1em} & CLIP (15m) & 43.4 & 72.3 & 33.8 & 44.2 & 15.8 & 2.3 & 14.0 & 53.8 &9.2&89.1 & 24.0 & 30.0 & 11.0 & 28.0 & 3.3 & 50.1 & 37.4 & 12.5 & 50.6 &10.2& 50.1 & 32.2  \\
& OC-CLIP (15m)  & 54.5 & 82.3 & 46.6 & 54.1 & 20.2 & 3.7 & 22.1 & 69.2 & 5.2&94.2 & 26.8 & 44.4 & 9.4 & 29.9 & 5.9 & 52.9 & 47.3 & 14.5& 51.3 &9.0& 49.9 & \textbf{45.0}  \\
\hline
\end{tabular}}

\end{adjustbox}
\caption{Zero-shot evaluation of CLIP vs OC-CLIP. Trained on varying size of data ( cc3m, cc12m, merged 15m) for 25 epochs.}
\label{tab:classification-downstream}

\end{table*}

\paragraph{Computational analysis of OC-CLIP} In OC-CLIP the visual and text modalities representations are no longer independent (as opposed to CLIP). A image representation is the results of some text-conditioned mechanism operated by the binding module. It essentially extracts relevant visual slots that constitutes the nodes of the scene graph coming from the caption. As a result, there is some notable computational overhead introduced by the additional cross-attention operations of the binding module. In particular : \begin{itemize}
    \item  1. The text encoder needs to encode the $N$ nodes and $R$ relations of the scene graph as opposed to a single sentence encoding in CLIP.
    \item 2. For each Image-Graph pair, The $N$ text nodes cross-attends to $N_{im}$ patches of the ViT in order to extract the structured visual slots.
\end{itemize}
When training OC-CLIP from scratch we propose to mitigate those two overheads respectively by : 
\begin{itemize}
    \item 1. Using a smaller embedding width (256 vs 512) and number of layers (6 vs 12) in the text encoder. Indeed OC-CLIP only need to encode information about objects and relationships and we expect such encoding to require much less capacity than an encoder that needs to encode a whole caption composed of multiple objects and relations between them.
    \item 2. We operate on a reduced embedding space 256 for the binding module and thus first project the ViT-B-16 patches from a 768 to a 256 embedding space before computing the nodes to patch cross attention logits.
    \item 3. We use the SigLip~\citep{zhai2023sigmoidlosslanguageimage} loss to make efficient use of batch chunking and gradient checkpointing.
\end{itemize}
We only perform experiments with a B-16 architecture for the ViT but perform the computational analysis fro both B and L backbones. We report the results in Table \ref{tab:vit-scale-binding-module} We note that there is  a significant overhead with a base architecture 2.2x but since the binding module perform the same number of operations no matter what the ViT is we show that when scaling the ViT backbone, the binding module is not the bottleneck anymore and the computational overhead is reduced (1.3x).
\begin{table}[htbp]
  \centering
  \begin{adjustbox}{width=\textwidth}
  \new{
  \begin{tabular}{|c|cc|ccc|c|}
    \hline
    Model & ViT Backbone & Text (w,l,ctx) & Binding Module GFLOPs  & Text GFLOPS & Vision GFLOPs & Total GFLOPs\\
    \hline
    OC-CLIP & B &(256, 6, 20)&12(*num workers) &180 & 1k& 2.2x \\
    CLIP & B & (512, 12, 77)& -& 186& 1k& 1x\\
    \hline
    OC-CLIP& L &(256, 6, 20)& 12(*num workers)& 180&4.9k& 1x \\
    CLIP& L&(512, 12, 77)& - & 186 &4.9k& 1.3x\\
    \hline
  \end{tabular}}
  \end{adjustbox}
  \caption{\new{Computational Comparison of CLIP and OC-CLIP. Calculations are made for a local batch size (per GPU) of 64. We give the Total GFLOPs based on a global batch size of 8192 (=128 num workers). When scaling the ViT backbone the computational overhead of the binding module remains fixed and is not the main bottleneck anymore.}}
  \label{tab:vit-scale-binding-module}
\end{table}
\subsection{Scene Graph Parsing Discussion}\label{app:parsing}
\paragraph{Comparison of different parsing methods} Although the parsing method is not the core of our contribution we provide here a couple of qualitative and quantitative comparisons to motivate the choice of using an LLM to perform the parsing of the captions despite the pre-processing computational overhead it entails. We identify 3 families of parsing method that operate on text-only input and provide insights on their respective : 
\begin{itemize}
    \item \textbf{Automatic parsing methods} : method based on hand-crafted rules about the semantics in order to extract tags and more complex dependency graphs. TagAlign also compares to nltk and justifies the choice of going to an llm-based method. We consider a representative of those automatic parsing methods based on spacy \citep{spacy2}.
    \item \textbf{Finetuned factual scene graph parser} trained in a supervised way to extract scene graph. We consider a representative of them, a state-of-the-art factual scene graph parser based on T5 model \citep{li2023factualbenchmarkfaithfulconsistent} trained to extract fine-grained scene graph information about the objects and relations in an input caption.
    \item \textbf{LLM-based}, here we choose llama3-8b as a representative and leave the extensive analysisof the bias/cues of different llm families of model for future work.
\end{itemize}
We identified failures modes of automatic parsing and finetuned that are relevant to compositional understanding of clip-like models and justify the use of an llm-based parsing method and summarize them in Table \ref{tab:parsing-errors}. We show on one hand that automatic parsing methods are prone to oversimplification, missing relations and mistaking an attribute modifiers with an object. On the other hand supervised scene graph parser seems to be prone to relation classification error and important atteibute binding error when the different objects mentioned in a caption share the same label tag.

\begin{table}[htbp]
  \centering
  \small
  \begin{adjustbox}{width=\textwidth}
  \new{
  \begin{tabular}{|l|l|l|l|}
    \hline
    \textbf{Caption} & \textbf{Spacy} & \textbf{T5} & \textbf{LLM} \\
    \hline
    A brown cat is lying on a computer & 
      \begin{tabular}{@{}l@{}}
        Objects: a brown cat, a computer \\
        Relations: \{\textcolor{red}{on}, 0, 1\} \quad \textcolor{red}{(Oversimplification error)}
      \end{tabular} &
      \begin{tabular}{@{}l@{}}
        Objects: brown cat, computer \\
        Relations: \{\textcolor{red}{lay on}, 0, 1\} \quad \textcolor{red}{(Relation classification error)}
      \end{tabular} &
      \begin{tabular}{@{}l@{}}
        Objects: brown cat, computer \\
        Relations: \{lying on, 0, 1\}
      \end{tabular} \\
    \hline
    A man is on the left of the dog & 
      \begin{tabular}{@{}l@{}}
        Objects: a man, \textcolor{red}{the left}, a dog \quad \textcolor{red}{(Wrong POS)} \\
        Relations: \{\textcolor{red}{of}, 1, 2\} \quad \textcolor{red}{(Missing relation)}
      \end{tabular} &
      \begin{tabular}{@{}l@{}}
        Objects: man, dog \\
        Relations: \{at the left of, 0, 1\}
      \end{tabular} &
      \begin{tabular}{@{}l@{}}
        Objects: man, dog \\
        Relations: \{on the left of, 0, 1\}
      \end{tabular} \\
    \hline
    A woman in blue and a woman in red & 
      \begin{tabular}{@{}l@{}}
        Objects: a woman, \textcolor{red}{red}, a woman,  \quad \textcolor{red}{(Wrong POS)} \\
        Relations: \{, 0, 1\}, \{in, 0, 2\}, in, 2, 3\} 
      \end{tabular} &
      \begin{tabular}{@{}l@{}}
        Objects: \textcolor{red}{blue red clothes}, woman \quad \textcolor{red}{(Wrong attribute binding)}\\
        Relations: \{wear, 0, 1\} 
      \end{tabular} &
      \begin{tabular}{@{}l@{}}
        Objects: woman in red, woman in blue \\
        Relations: \{\}
      \end{tabular} \\
    \hline
  \end{tabular}}
  \end{adjustbox}
  \caption{\new{Comparison of parsing errors made by different parsers.}}
  \label{tab:parsing-errors}
\end{table}

We additionally train OC-CLIP on COCO captions parsed by those 3 different parsing models and compare the downstream compositional understanding performance in Figure \ref{fig:parsing-perf}. Coherent with the qualitative analysis the choice of the parsing family mostly impact relational understanding. We observe for the SugarCrepe swap object (replace rel resp.)  a decrease of $9.3\%$ (resp. $14.1\%$) for spacy and $3.4\%$ (resp. $6.3\%$) for a supervised T5 model  as compared to OC-CLIP on scene graphs extracted by llama3-8b. 
Close to our work, TagAlign\citep{liu2024tagalignimprovingvisionlanguagealignment} also quantitatively and qualitatively analyze the objects tags than can be extracted with an nltk-based and llm-based parser and show that training CLIP with an additional object and attribute tag classification loss with tags coming from an llm results in better downstream zero-shot semantic segemntation.
\begin{figure}
    \centering
    \includegraphics[width=\linewidth]{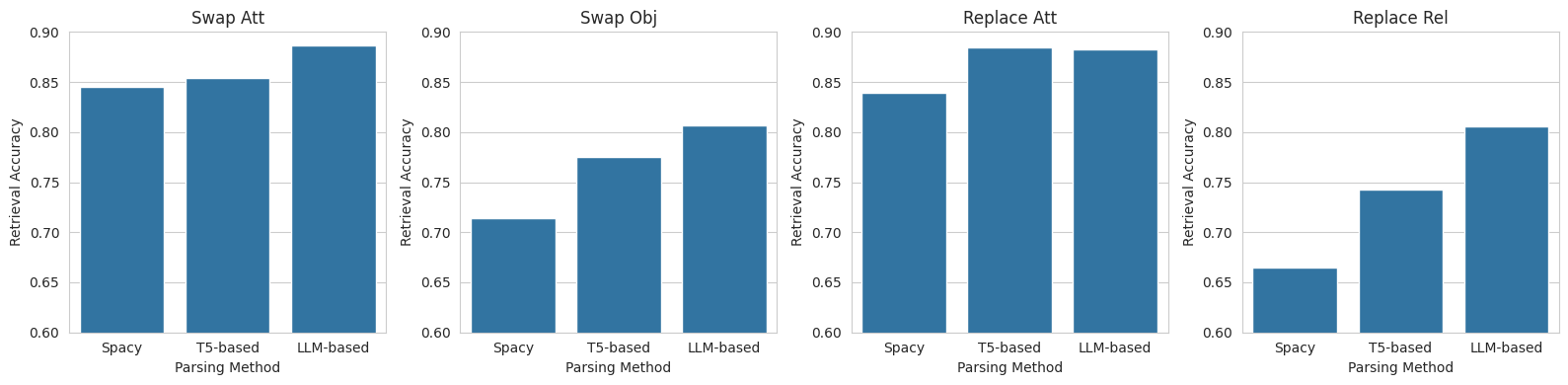}
    \caption{\new{Downstream Compositional Understanding of OC-CLIP when trained on different parsing of COCO-Captions.}}
    \label{fig:parsing-perf}
\end{figure}

\paragraph{Limitations of LLM-based parsing for OC-CLIP}
We also acknowledge that using and LLM as a parser may also have some limitations and evaluating the impact of the downstream performance of different LLMs or VLMs is an interesting question. In particular, llm-based parsing might not extract accurate scene graphs, especially when the dependency between the objects in a captions is rather complex or ambiguous. And informing the parser in prompt with visual information might be an interesting direction. However the exact instanciation of the LLM-based parser used is orthogonal to our contribution and we leave this analysis for future work.

\paragraph{Scene Graph Parsing cost}
We performed the parsing by serving instances of Llama3.1-8b on v100 GPUs. Each datasets is then chunked in $N$ process that do not require any GPUs and send requests to the served LLM parsers through vllm\footnote{https://github.com/vllm-project/vllm} to maximize the throughput of the parallelized requests. For reference we parsed the COCO datasets ($\sim$ 500k captions) parallelizing 10 instances of the parser, and with 128 chunks in 3.5 hours and Visual-Genome ($\sim$ 200k captions) with 8 instances, 64 chunks in 1.7hours. The parsing time can further be optimized by serving more instances, using more performant GPUs (A100, H100 etc..), serving each instance in parallel in more GPUs to maximized the number of requests that can be processed per second.
For the cc3m and cc12m, in order to accelerate the parsing, we kept the LLM parser local using ollama\footnote{https://ollama.com/} on v100 GPUs. CC3M was chunked into 500 and CC12M into 1000 smaller chunks, we launched the jobs sequentially. CC12M took about ~3days to parse but could likewise be accelerated using faster GPUS.

}

\subsection{Additional Compositional Understanding results}\label{app:vl}
Our main goal is to evaluate CLIP-like models compositional understanding in plausible and grammatically correct cases. \citet{hsieh2023sugarcrepefixinghackablebenchmarks} have identified exploitable textual biases in previous mainstream procedurally-generated hard negatives benchmarks like the COCO and Flickr set of ARO and VL-checklist. Specifically they show that procedurally generated hard negatives are either highly grammatically incorrect and can be identified by a blind model or by a good language model that can measure the plausibility of the caption. The SugarCrepe is thus designed to  pfollow the same fine-grained taxonomy on attributes, objects, relationships as VL-checklist but ensures that the hard-negative are not distinguishable by a blind model. The main results of our paper thus focus on this benchmark. We however also give the performance of our model on the full ARO suite and VL-Checklist in Table~\ref{tab:vl_aro} for reference.  
\begin{table}[htbp]
\begin{small}
\begin{sc}
\resizebox{\textwidth}{!}{
\begin{tabular}{lcccccccc}

\toprule
Model & \multicolumn{3}{c}{VL-Checklist} & \multicolumn{4}{c}{ARO} \\
\cmidrule(lr){2-4}
\cmidrule(lr){5-8} 
& Object & Relation   & Attribute & Attribution & Relation & COCO-order & Flickr-order \\
\midrule
CLIP & 80.0 & 63.0&  67.4& 63.2& 60.0 & 47.9 & 60.2 \\
BLIP & 82.2 & 70.5 &  75.2& 63.2& 60.0 & 47.9 & 60.2 \\
XVLM & 85.8 & 70.4 &  75.1& 73.4& 86.8 & - & -\\
\midrule
\textit{Hard-negative Methods} \\
CLIP-SVLC &85.0 & 68.9.7 & 72.0 & 73.0 & 80.6 & 84.7 & 91.7 \\
NegCLIP& 84.1& 63.5& 70.9 & 71 & 81 & 86 & 91 \\
CE-CLIP& 84.6 &71.8 &  72.6& 76.4 & 83.0 & - & - \\
\textit{Dense captioning+Hard-Negative} \\
DAC-LLM$_\text{500k}$ & 66.5 & 56.8 & 57.4 & 63.8 & 60.1 & 50.2 & 61.6 \\
DAC-LLM$_\text{3M}$ & 87.3 & 86.4  & 77.3 & 73.9 & 81.3 & 94.5 & 95.7 \\
DAC-SAM$_\text{3M}$ & 88.5 & 89.7 & 75.8 & 70.5 & 77.2 & 91.2 & 93.9 \\  
DCI & 80.7& 70.1 & 68.7 & 67.6 & 76.2 & 88.6 & 91.3 \\
DCI$_\text{neg}$ & 88.4 & 61.3 & 70.4 & 62.0 & 57.3 & 39.4 & 44.6\\
\midrule
OC-CLIP & 90.7 & 80.0 & 75.6 & 84.0 & 84.9 & 94.2 & 84.8 \\
\bottomrule
\end{tabular}
}
\end{sc}
\end{small}

\caption{\new{Results (\%) on VL-Checklist  and ARO Benchmark.}}
\label{tab:vl_aro}
\end{table}
\subsection{PUG Dataset }\label{sec:pug_app}
In this section we describe in more details the content of the synthetic experiments, give more context on the motivation along with additional results. 
\paragraph{Motivation}The rise of data-centric hard negative methods were motivated by the bag-of-words behaviour \citep{yuksekgonul2023when} of CLIP noticed in "simple swap-attribute" retrieval tasks. Hard-negative methods propose to mitigate this behaviour by finetuning CLIP-like models  on data points with minimal changes but semantically different meanings. However we experimentally observed that all the methods fail to increase performance specifically in swap attribute kind of splits. In order to further isolate the root cause, we propose a series of synthetic experiments that compare covering more hard-negative data points with OC-CLIP on varying proportion of training samples and hard-negative samples. By restricting the environment to a closed-set vocabulary of backgrounds, attributes, and object classes, we can enumerate all possible hard-negatives, allowing us to systematically evaluate the effectiveness of different approaches.
Our results show that simply \emph{adding more hard-negatives plateaus and is not sample-efficient, as the swap attribute binding performance always underperforms OC-CLIP trained on less data without any hard-negatives} in a simple object-attribute binding task \ref{fig:binding_pug}. However, when combined with OC-CLIP inductive bias, hard-negatives complementarily improve downstream performance. This suggests that our model, OC-CLIP, is a more sample-efficient approach to addressing the bag-of-words behavior of CLIP models.
We hypothesize that the root cause of this issue thus lies in the representation format used in CLIP's original formulation, which relies on a single vector to capture complex semantic relationships. Our proposed method introduces inductive biases that allow the model to learn more structured representations, avoiding superposition of features \citep{greff2020bindingproblemartificialneural} and effectively mitigating the bag-of-words behavior. Through these synthetic experiments, we demonstrate the effectiveness of our approach and provide insights into the sample-efficiency limitations of existing data-centric methods.

\begin{figure*}[ht]
\centering
    \begin{subfigure}[b]{0.15\textwidth}
        \centering    \includegraphics[width=0.9\textwidth, keepaspectratio]{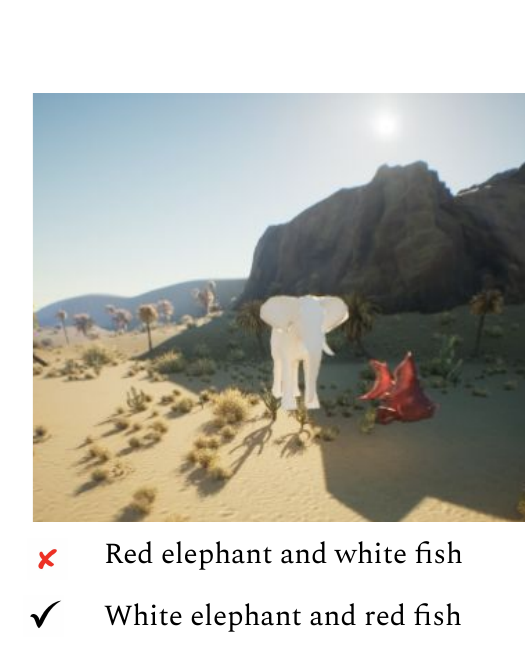}
        \caption{\tiny{} Pairs  }
    \end{subfigure}%
    \begin{subfigure}[b]{0.2\textwidth}
        \centering    \includegraphics[width=0.9\textwidth,  keepaspectratio]{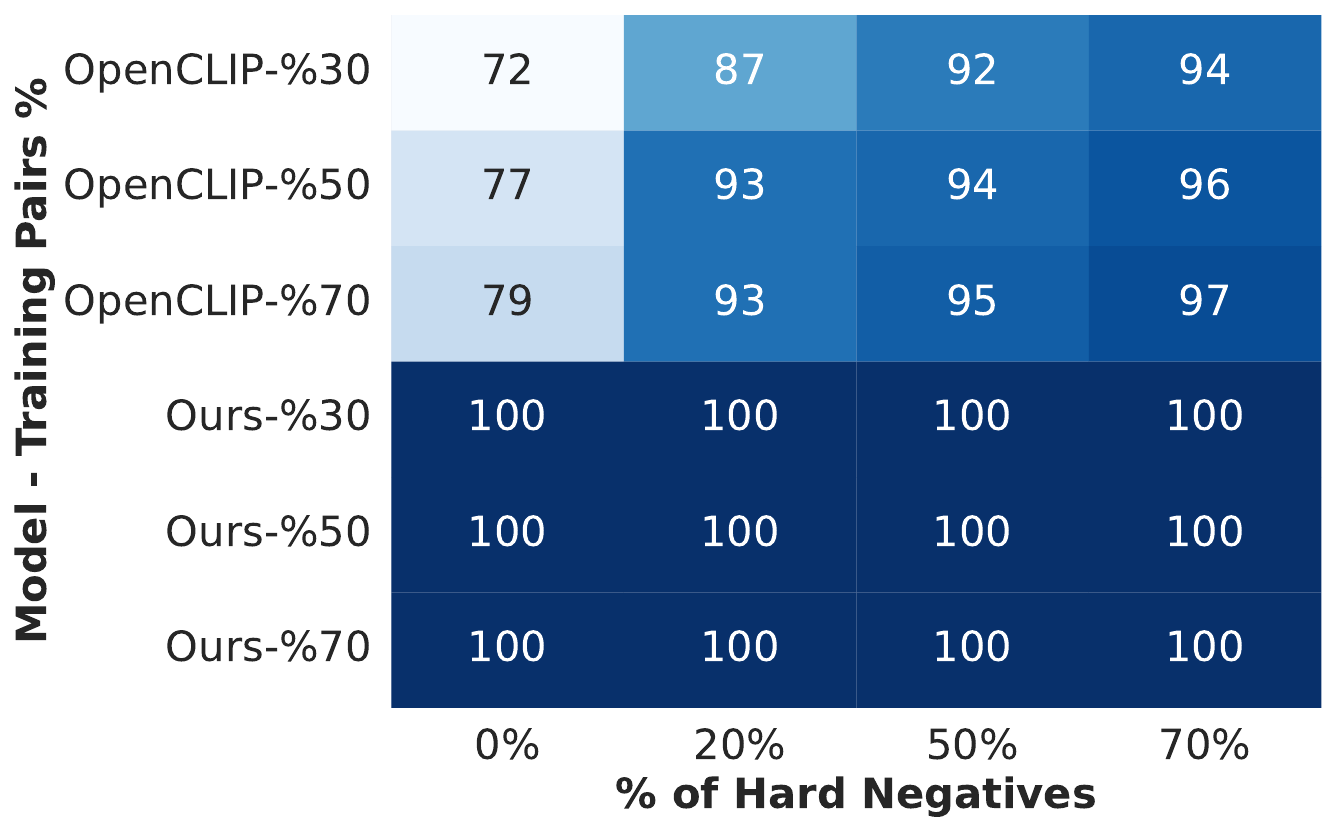}
        \caption{\tiny{} Train Pairs}
    \end{subfigure}%
    \hfill
    \begin{subfigure}[b]{0.2\textwidth}
        \centering    \includegraphics[width=0.9\textwidth,  keepaspectratio]{figures/pairs_1.pdf}
        \caption{\tiny{} Seen Pairs}
    \end{subfigure}%
    \hfill
    \begin{subfigure}[b]{0.2\textwidth}
        \centering    \includegraphics[width=0.9\textwidth,  keepaspectratio]{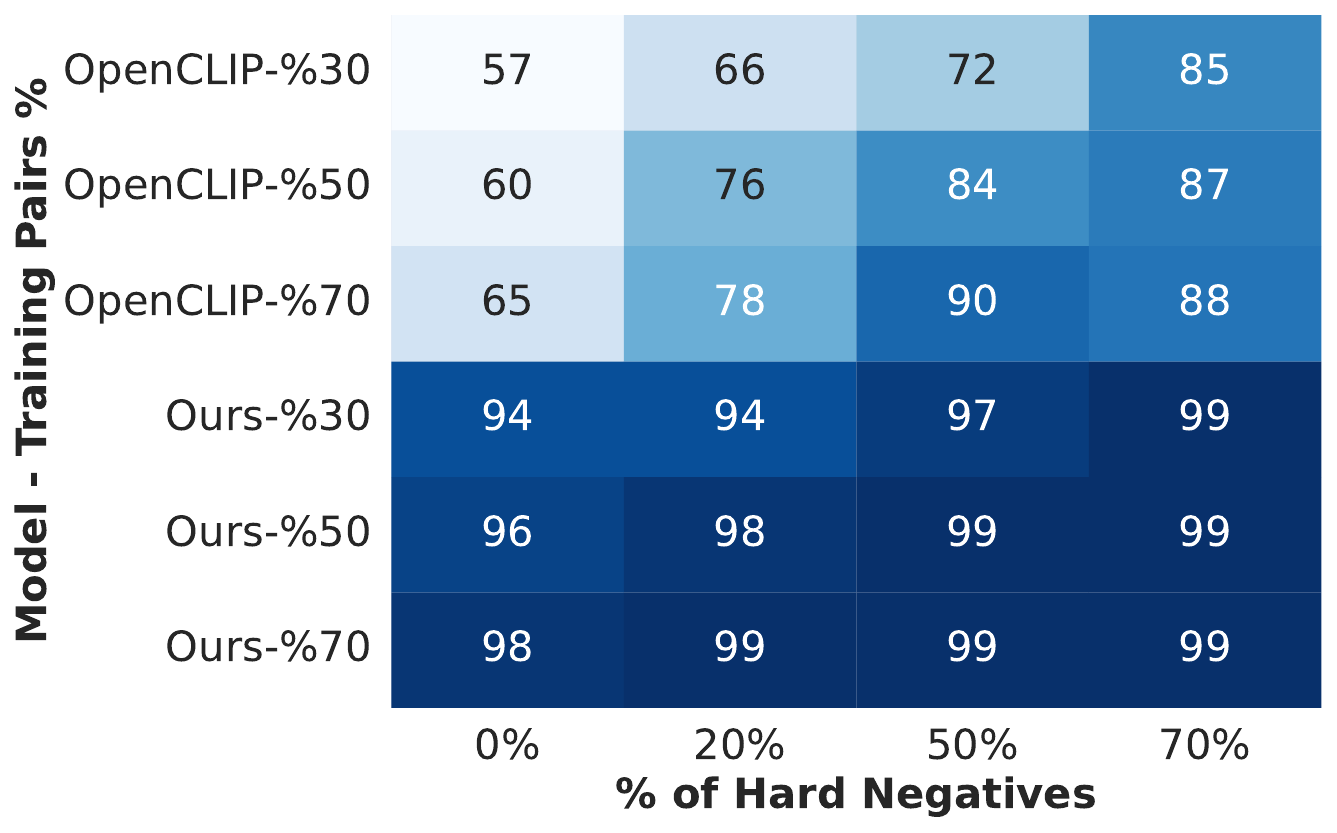}
        \caption{\tiny{} Different Bag-of-words}
    \end{subfigure}%
    \hfill
    \begin{subfigure}[b]{0.2\textwidth}
        \centering    \includegraphics[width=0.9\textwidth, keepaspectratio]{figures/pairs_3.pdf}
        \caption{\tiny{} Unseen Pairs}
    \end{subfigure}%

    \begin{subfigure}[b]{0.15\textwidth}
        \centering    \includegraphics[width=\textwidth,  keepaspectratio]{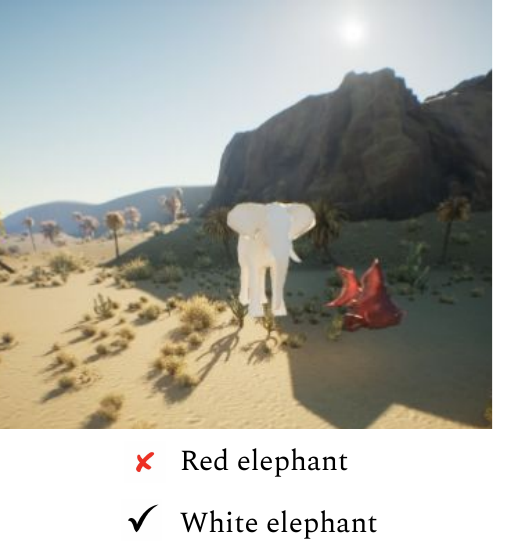}
        \caption{\tiny{} Single Object }
    \end{subfigure}%
    \begin{subfigure}[b]{0.2\textwidth}
        \centering    \includegraphics[width=\textwidth, keepaspectratio]{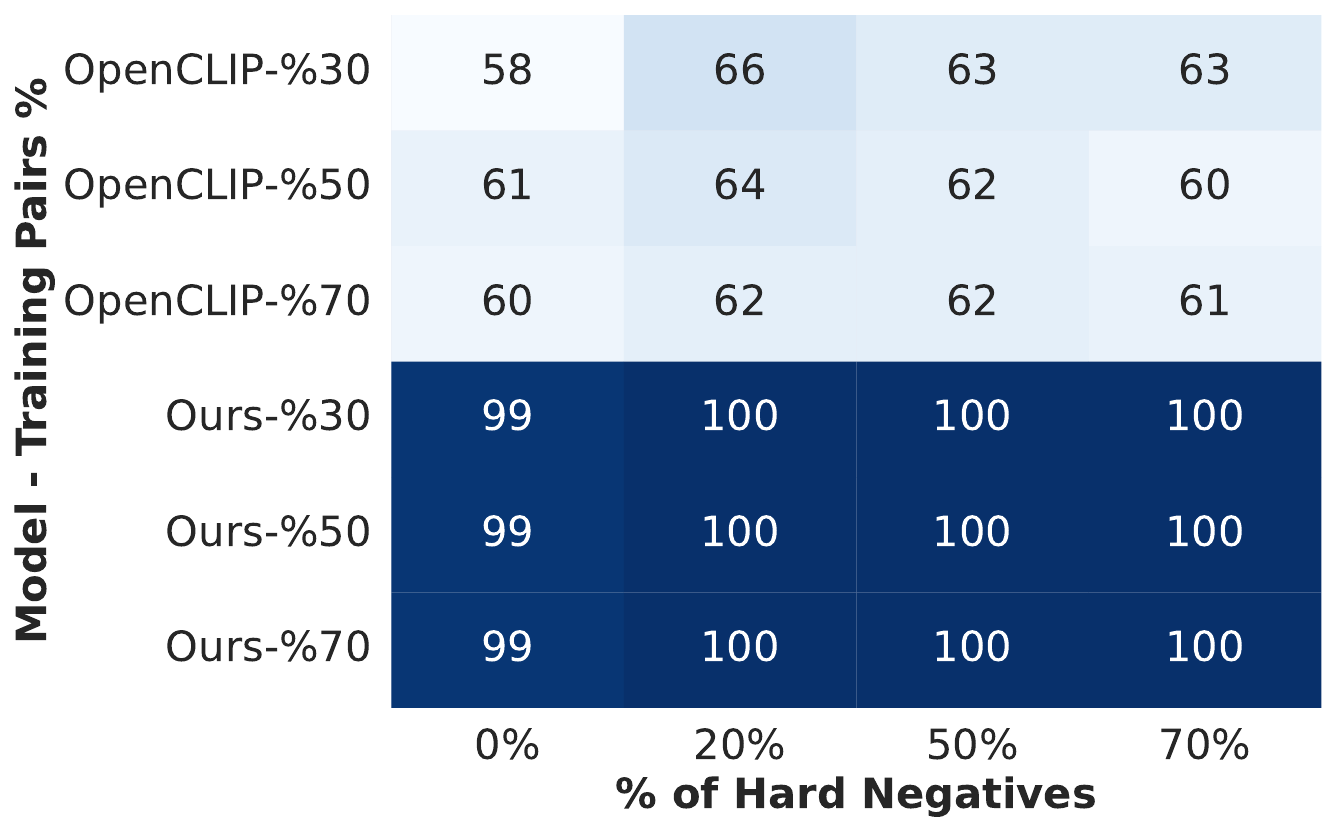}
        \caption{\tiny{} Train Pairs}
    \end{subfigure}%
    \hfill
    \begin{subfigure}[b]{0.2\textwidth}
        \centering    \includegraphics[width=\textwidth, keepaspectratio]{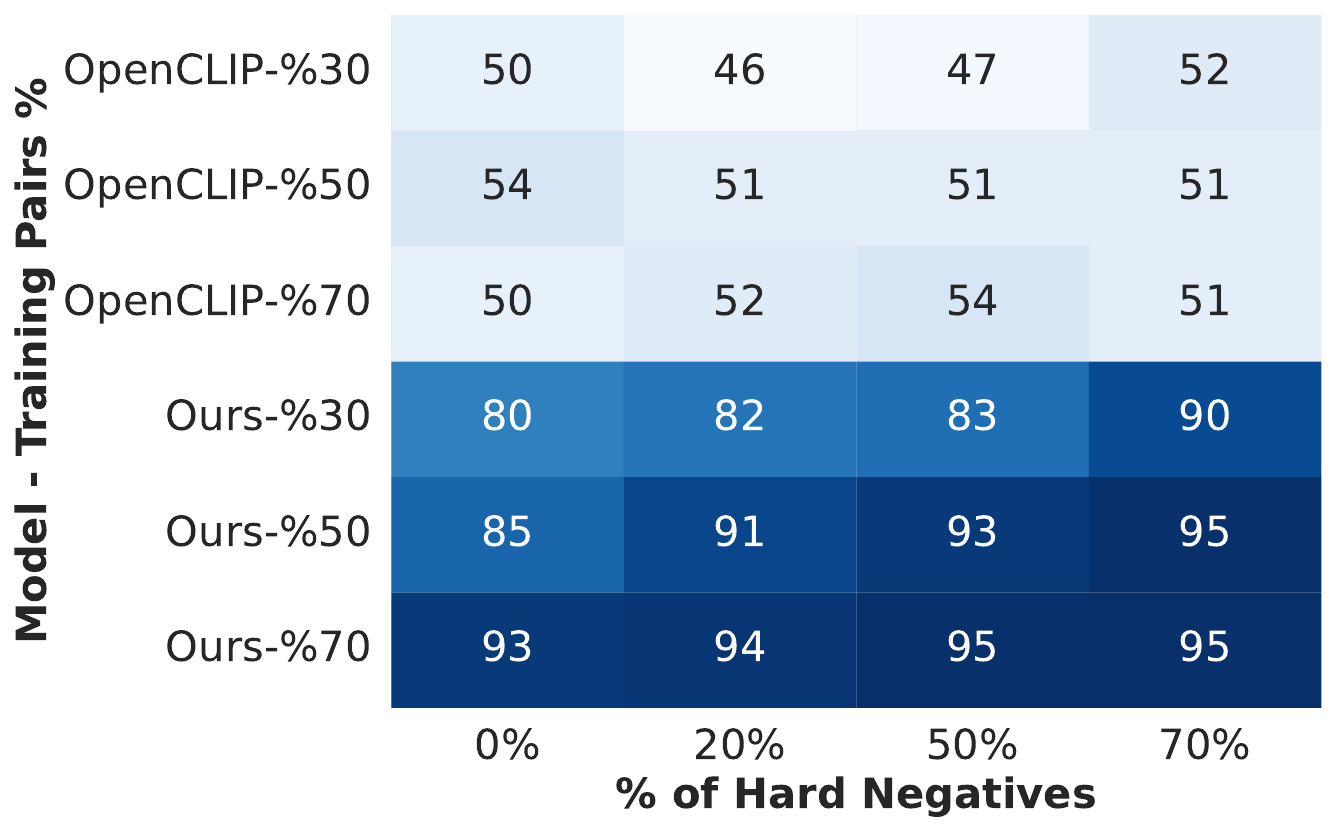}
        \caption{\tiny{} Seen Pairs}
    \end{subfigure}%
    \hfill
    \begin{subfigure}[b]{0.2\textwidth}
        \centering    \includegraphics[width=\textwidth,  keepaspectratio]{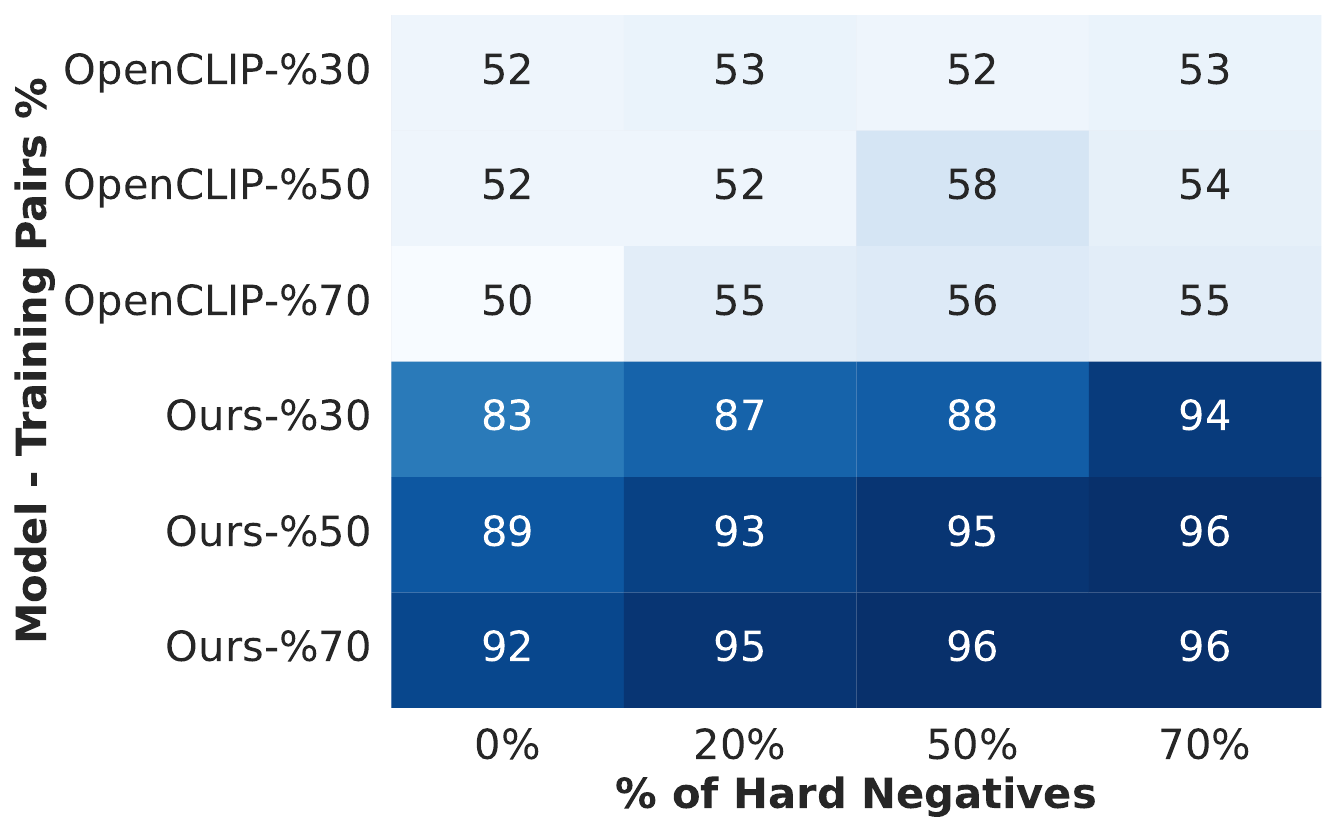}
        \caption{\tiny{} Different Bag-of-words}
    \end{subfigure}%
    \hfill
    \begin{subfigure}[b]{0.2\textwidth}
        \centering    \includegraphics[width=\textwidth,  keepaspectratio]{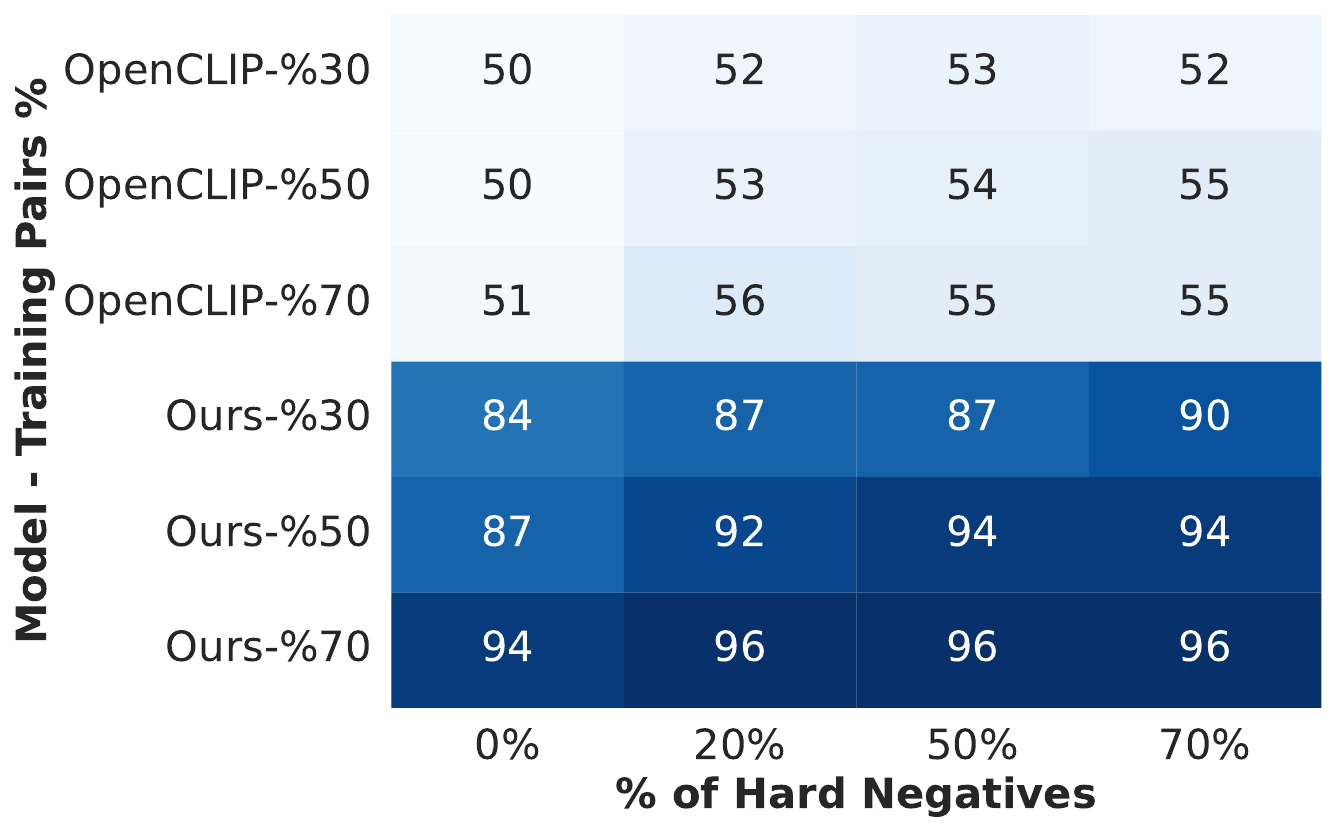}
        \caption{\tiny{} Unseen Pairs}
    \end{subfigure}%

  \caption{\small{\textbf{Attribute Binding on PUG - Additional Results}
  Performance of the finetuned OpenCLIP and OC-CLIP models on a binary classification task between a caption and its corresponding hard-negative. We do that for captions that mention Pairs of animals (\textbf{top row}) like the example in Figure (a) and for captions that mention a single animal (\textbf{bottom row}) like the example in Figure (b).To assess the models' performance, we compute the accuracy across two dimensions. The first one is the percentage of animal pairs (y-axis) seen during training (animals like elephants and fish could be seen either alone or with other animals but never together). The second dimension (x-axis) is the number of hard-negatives used in the training data. For instance, whether we have the combination ``red elephant'' and ``white fish'' in the training data while we only have ``white elephant'' and ``red fish'' in the test data. }}
  \label{fig:abla_attr}
\end{figure*}

\paragraph{Dataset splits} The synthetic experiments we propose are based on  the controlled 3D environment  PUG~\citep{bordes2023pugphotorealisticsemanticallycontrollable}. We operate in a 3D envrionment with pairs or single textured animals in different backgrounds. The factors of variation are :
\begin{itemize}
    \item \textbf{5 Backgrounds} : desert, arena, ocean floor, city, circus
    \item \textbf{20  Animals}  : goldfish, caribou, elephant, camel, penguin, zebra,
        bear, crocodile, armadillo, cat, gecko, crow,
        gianttortoise, rhinoceros, dolphin, lion, orca, pig,
        rabbit, squirrel
    \item \textbf{4 textures} : red, white, asphalt, grass
    \item \textbf{2 spatial constraints} for pairs : left/right, above/under
\end{itemize}
The different splits 
We then construct splits that aim at evaluating separately attribute binding and spatial relationships understanding. In all the different splits, we include images with single animals in all the possible \emph{background-texture-animal} conjunctions.
\paragraph{Attribute Binding Splits} The attribute binding training and testing splits are constructed as follows : (1) - We list all the possible pairs of animals,(2) - We randomly  and i.i.d. select a percentage \%  $N_{\text{train}}$ of pairs to include in the train split, (3) - For each training pair we select a pair of assigned attribute (for example if cat and caribou are in the train split we will assign red to cat and white to caribou and will remove all the other attribute-animal conjunction from the training. This is done such that we can control for the \emph{replace attribute} hard negative presence. (4) - For each pair in the training set we separate the corresponding hard negative examples with the same bag of words but swapped attributes (referred to as \emph{Seen Pairs} in Figure \ref{fig:abla_attr}) and the same pair but a different bag of words ( referred to as \emph{Different Bag-of-words} in \ref{fig:abla_attr})  , (5) - finally we also isolate unseen pairs of animals. We also include the accuracy on the training pairs that do not have their corresponding hard negatives in the test set).

\paragraph{Spatial Relation understanding Splits} For these splits we do not assign specific pairs of attributes to train/test split but rather consider pairs of animals and their order with respect to the spatial relationship tested and systematically include all the possible attributes assignment to those pairs. We then construct the different splits by restricting the number of pairs and their spatial configuration. 
\paragraph{Hard Negative Samples} For both tasks the hard negative samples we consider are align with the test tasks taxonomy. For attribute binding we always test the model's ability to distinguish between eg. \emph{a red cat and a white caribou} and \emph{a white cat and a red caribou}. Hence we consider as a hard negative sample any image that corresponds to the swapped attribute version of a training pairs. To augment the dataset with hard negative, we sample i.i.d. a percentage \%  $N_{\text{hard}}$ of the training pairs and include in their corresponding hard negatives in the train set.
Similarly for the spatial relationship understanding task, we test the model's ability to distinguish between eg. \emph{a red cat to the left of a white caribou} and \emph{a white caribou to the left of a red cat}. Hence we consider as a hard negative sample any image that corresponds to the swapped order with respect to the relationship tested of the animal pairs seen during training.
    
\subsubsection{Spatial Relation Understanding}
\label{PUG-rel}
In this section, we aim to evaluate the spatial relationship understanding capabilities of the models. 
To do so, we conduct controlled experiments using data splits where not all pairs of animals are seen during training. The relations considered in these experiments are ``left/right'' and ``above/below''. Hence, the task is to choose between the original caption of the form ``{X} left of {Y}'' and the caption with the swapped order ``{Y} left of {X}''. We consider the following generalization axes:\looseness-1
\begin{itemize}
    \item \textbf{Unseen object order:} This axis tests the generalization when swapping the order of objects in a relationship. For example, ``elephant to the left of fish'' may be used for training, while ``elephant to the right of fish'' is used for evaluation
    \item \textbf{Unseen object pairs:} This axis test for unseen pairs of animals in seen relationships.\looseness-1
\end{itemize}
We follow the experimental setup of section~\ref{PUG-attr}, and finetune OpenCLIP and OC-CLIP while considering the effect of adding different \% of hard negative images and/or different \% of object pairs to the training data.\looseness-1

\begin{figure*}[ht]
\centering
    \begin{subfigure}[b]{0.33\textwidth}
        \centering    \includegraphics[width=0.8\textwidth]{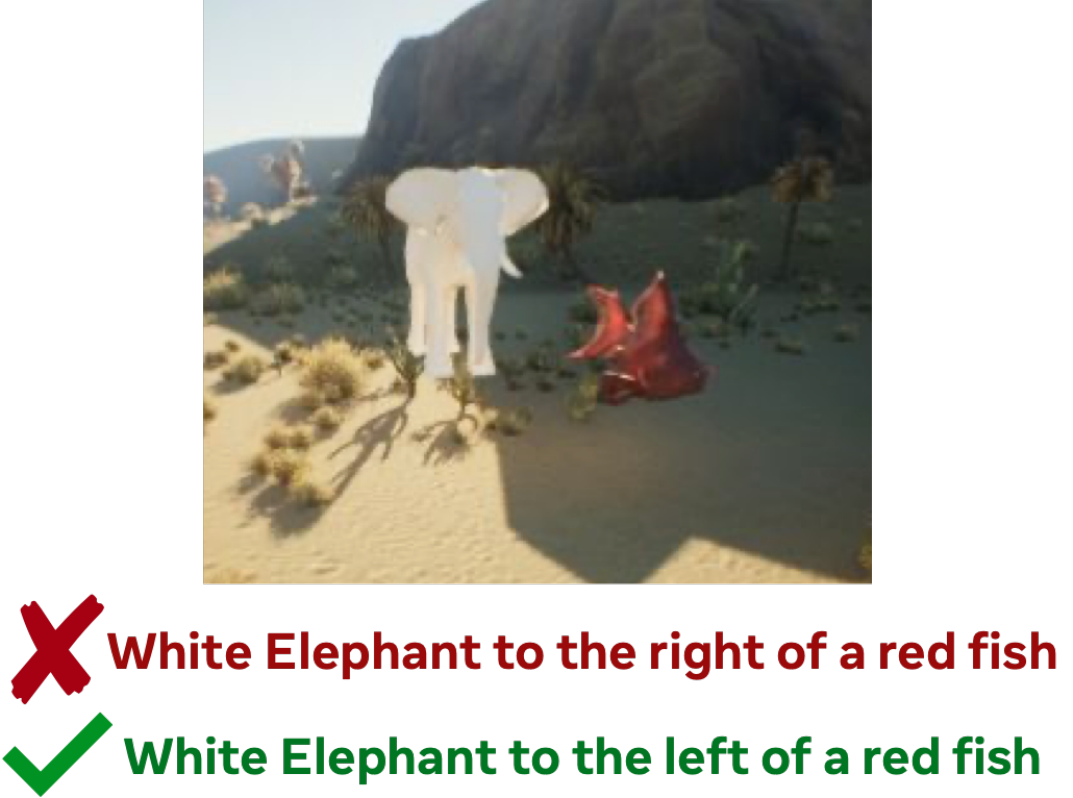}
        \caption{ Example image}
    \end{subfigure}%
    \hfill
    \begin{subfigure}[b]{0.33\textwidth}
        \centering    \includegraphics[height=1.13in]{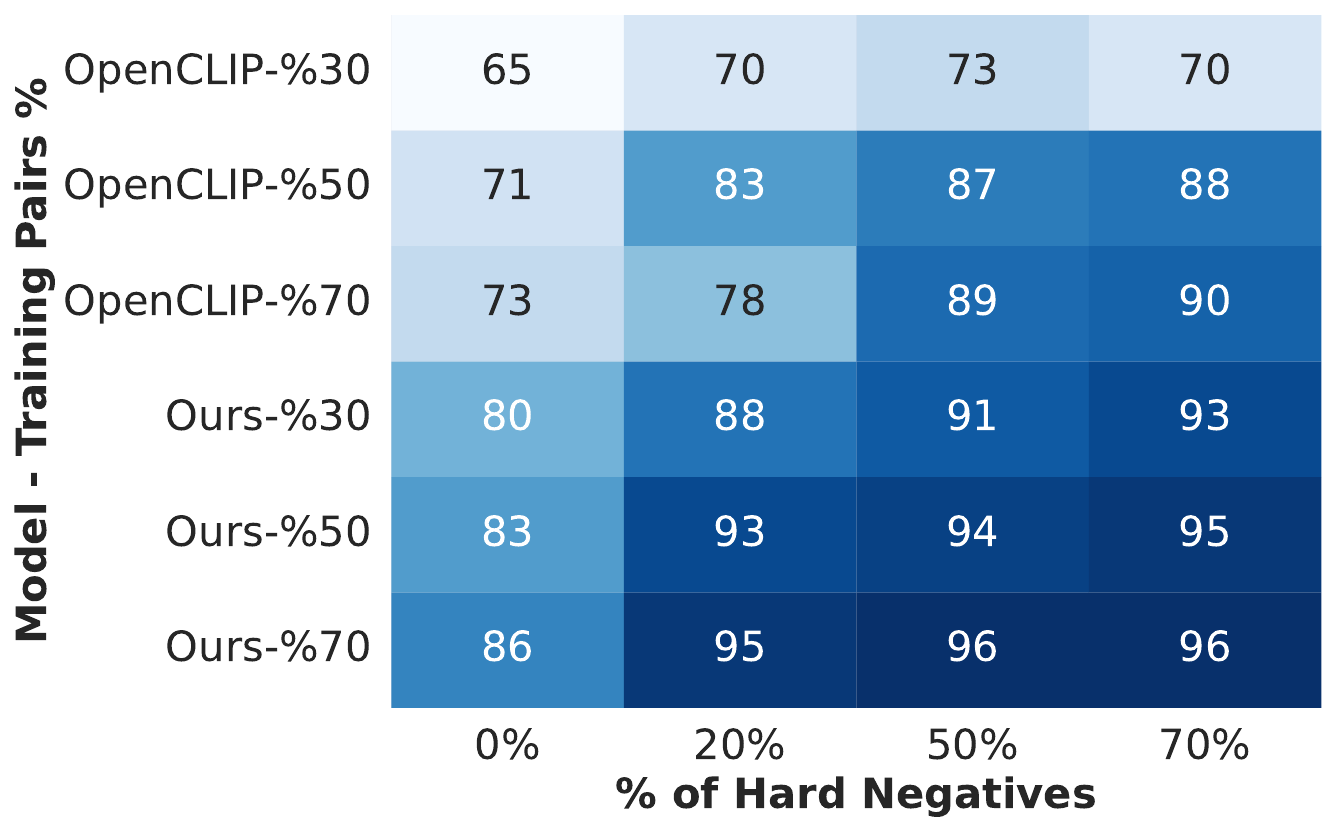}
        \caption{Unseen object order}
    \end{subfigure}%
    \begin{subfigure}[b]{0.33\textwidth}
        \centering    \includegraphics[height=1.13in]{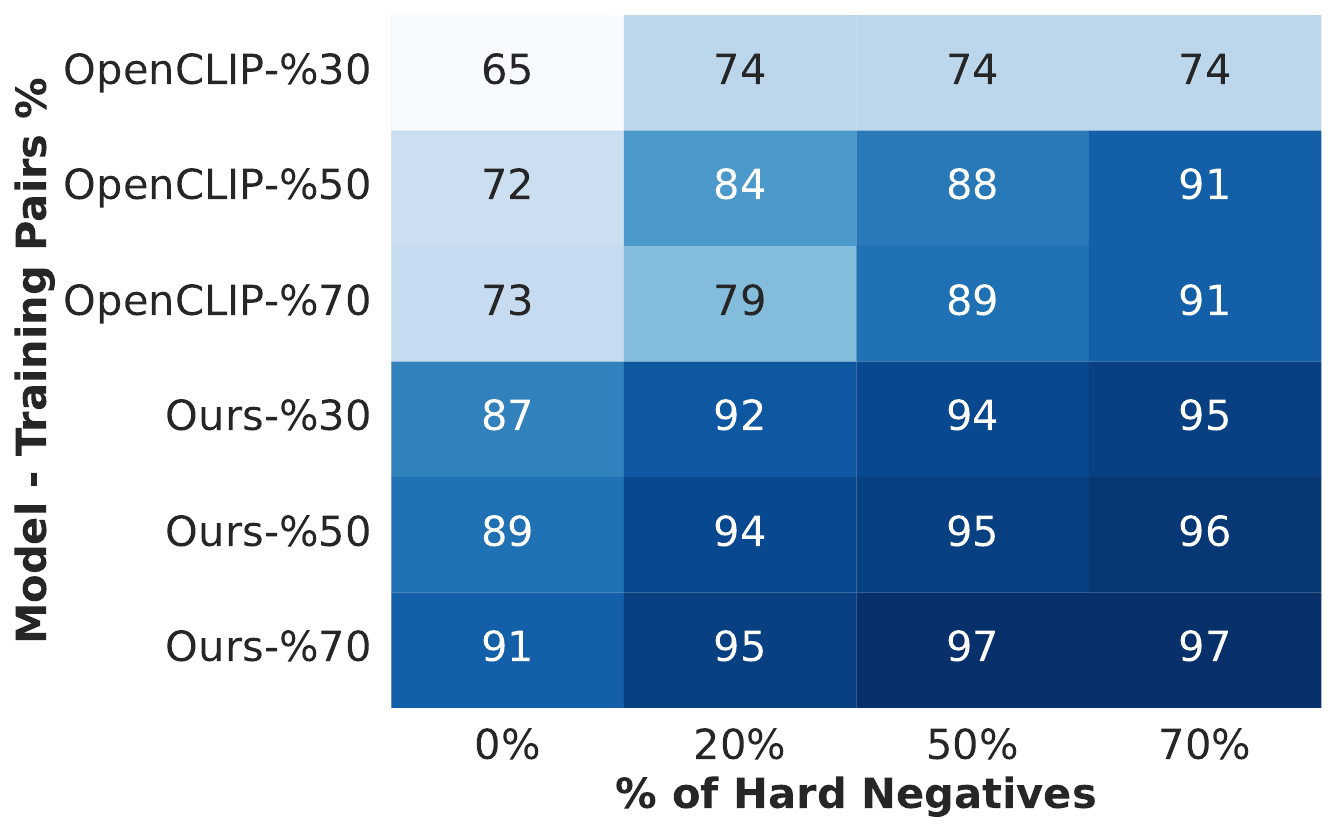}
        \caption{Unseen object pairs}
    \end{subfigure}

  \caption{\textbf{Spatial Relationship Understanding.} We finetune OpenCLIP and train OC-CLIP's binding module on splits containing different \% of animals pairs (y-axis) and different \% of hard-negative image in the training split (x- axis). We test the models on images with either unseen order (b) or unseen pairs (c) during training. The testing is done  against the swapped order of the ground truth caption as shown in the visual example (a).\looseness-1}
  \label{fig:rel_pug}
\end{figure*}

We test both models on image-text retrieval tasks and report the results in Figure~\ref{fig:rel_pug}. Figure~\ref{fig:rel_pug}(b) shows the results for the unseen object order generalization, whereas Figure~\ref{fig:rel_pug}(c) presents the results for the unseen object pairs. As shown in Figure~\ref{fig:rel_pug}(b), OC-CLIP outperforms OpenCLIP in all data regimes considered, with improvements between $6\%$ and $18\%$. Similarly, as shown in Figure~\ref{fig:rel_pug}(c), OC-CLIP improves upon OpenCLIP in all data regimes, yielding absolute improvements between $5\%$ and $20\%$.\looseness-1

\subsection{Parsing}

For the parsing of the training and testing data we used a llama-3-70b Instruct model with the following prompt :
\vspace{-1em}
\begin{tcolorbox}[colback=gray!5!white, colframe=gray!75!black, title=Parsing Prompt]
\footnotesize
Given a caption, your task is to parse it into its constituent noun phrases and relationships. The noun phrases should represent independent visual objects mentioned in the caption without semantic oversimplification.
For each caption, output the parsed noun phrases (e.g., entities) and relationships in JSON format, placing the dictionary between \texttt{[ANS]} and \texttt{[/ANS]} brackets. In the relationships, use indices to specify the subject and object of the relationship mentioned in the caption. The indices of the subject and object should be integers.
Here are a few examples:
\begin{lstlisting}
Caption: A large brown box with a green toy in it
Output:
[ANS]
{
  "entities": [
    "large brown box",
    "green toy"
  ],
  "relationships": [
    {
      "relationship": "in",
      "subject": 1,
      "object": 0
    }
  ]
}
[/ANS]

[...]  More examples
\end{lstlisting}

PAY ATTENTION to the following:\\
- Relationships MUST relate two different entities in the caption and NOT be unary. For example, in the caption 'red suitcases stacked upon each other', 'stacked upon each other' is not considered a relationship.\\
- Do not forget any relationships.\\
- Relationships MUST be directed. 'and' is not a relationship.\\
- Pay attention to spatial relationships like 'behind', 'left of', 'with', 'below', 'next to', etc. 'and' is not a relationship.\\
- Check the right dependencies when the relationships are not direct. In the caption template a X with a Y in it, it refers to X.\\
- Pay attention to co-references.\\ \\
Now, parse the following caption into its constituting entities and relationships. You MUST place the answer between \texttt{[ANS]} and \texttt{[/ANS]} delimiters.\\
Caption:
\end{tcolorbox}

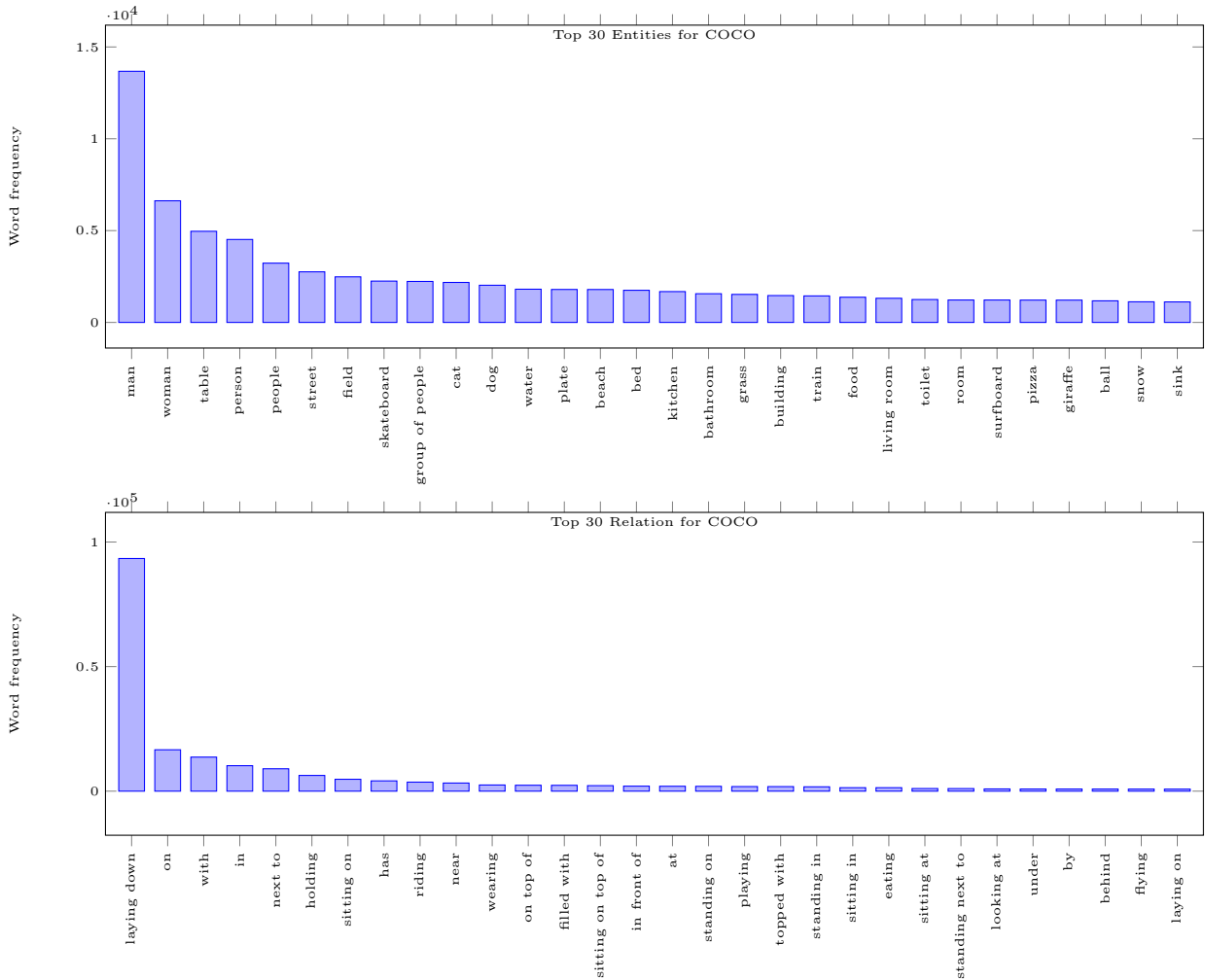
\begin{figure}[ht]
  \centering
\begin{tikzpicture}
\tiny
  \begin{axis}[
    xtick=data,
    ybar,
    xlabel=Top 30 Entities for COCO,
    ylabel=Word frequency,
    width=\textwidth,
    height=6cm,
    enlarge x limits=0.025,
    enlarge y limits=0.2,
    xticklabel style={rotate=90},
    xlabel style={at={(0.5,1.05)}, anchor=south},
    symbolic x coords={man,woman,table,person,people,street,field,skateboard,group of people,cat,dog,water,plate,beach,bed,kitchen,bathroom,grass,building,train,food,living room,toilet,room,surfboard,pizza,giraffe,ball,snow,sink,}
  ]
  \addplot coordinates {
    (man, 13676)
    (woman, 6624)
    (table, 4963)
    (person, 4518)
    (people, 3228)
    (street, 2756)
    (field, 2482)
    (skateboard, 2251)
    (group of people, 2233)
    (cat, 2178)
    (dog, 2025)
    (water, 1809)
    (plate, 1795)
    (beach, 1792)
    (bed, 1750)
    (kitchen, 1679)
    (bathroom, 1563)
    (grass, 1526)
    (building, 1463)
    (train, 1444)
    (food, 1372)
    (living room, 1315)
    (toilet, 1247)
    (room, 1222)
    (surfboard, 1221)
    (pizza, 1216)
    (giraffe, 1215)
    (ball, 1176)
    (snow, 1123)
    (sink, 1118)
  };
  \end{axis}
\end{tikzpicture}
\begin{tikzpicture}
\tiny
  \begin{axis}[
    xtick=data,
    ybar,
    xlabel=Top 30 Relation for COCO,
    ylabel=Word frequency,
    width=\textwidth,
    height=6cm,
    enlarge x limits=0.025,
    enlarge y limits=0.2,
    xticklabel style={rotate=90},
    xlabel style={at={(0.5,1.05)}, anchor=south},
    symbolic x coords={laying down,on,with,in,next to,holding,sitting on,has,riding,near,wearing,on top of,filled with,sitting on top of,in front of,at,standing on,playing,topped with,standing in,sitting in,eating,sitting at,standing next to,looking at,under,by,behind,flying,laying on,}
  ]
  \addplot coordinates {
    (laying down, 93385)
    (on, 16607)
    (with, 13665)
    (in, 10224)
    (next to, 8983)
    (holding, 6307)
    (sitting on, 4772)
    (has, 4126)
    (riding, 3618)
    (near, 3254)
    (wearing, 2474)
    (on top of, 2374)
    (filled with, 2344)
    (sitting on top of, 2206)
    (in front of, 2007)
    (at, 1962)
    (standing on, 1932)
    (playing, 1832)
    (topped with, 1820)
    (standing in, 1740)
    (sitting in, 1391)
    (eating, 1390)
    (sitting at, 1052)
    (standing next to, 1031)
    (looking at, 890)
    (under, 860)
    (by, 855)
    (behind, 849)
    (flying, 831)
    (laying on, 817)
  };
  \end{axis};
\end{tikzpicture}
 \caption{Plot of the most common entities and relations that were extracted by our LLM-based parser for the COCO datasets.}
 \label{fig:coco_most_common}
\end{figure}

To showcase the effectiveness of this parser, we are showcasing below the most and least common entities and relations that are find by this parser across MS-COCO and CC12M. In Figure \ref{fig:coco_most_common}, we can see that the most common entities are people for the COCO dataset as well as for CC12M in Figure \ref{fig:cc12m_most_common}. We also plot the least common entities in Figure \ref{fig:coco_least_common}. Finally, we also compute the number of tokens that is require for modelling the entities and relations. In Figure \ref{fig:coco_token_dist}, we display the frequency of the number of tokens use to encode the relations and entities on COCO. Using just 10 tokens, we can encode most of the entities and relations, thus we do not need to have a text encoder that take into input 77 tokens but can use a much smaller one instead. In Figure \ref{fig:cc12m_token_dist}, we show a similar plot but normalized. Even on a dataset with noisy captions like CC12M, most entities and relations can be encoded with less than 20 tokens.

\begin{figure}[ht]
  \centering
\begin{tikzpicture}
\tiny
  \begin{axis}[
    xtick=data,
    ybar,
    xlabel=Least 30 Entities for COCO,
    ylabel=Word frequency,
    width=\textwidth,
    height=6cm,
    enlarge x limits=0.025,
    enlarge y limits=0.2,
    xticklabel style={rotate=90},
    xlabel style={at={(0.5,1.05)}, anchor=south},
    symbolic x coords={abandoned,delapidated,light colored countertop,asphalt lot,back of a refrigerator,plastic portable computer,styrofoam tray,right hip,group of three boys,flag bearing motorcycles,Clive,funky looking purple scissors,but,white palm pilot,chicken burger,plug,vandalized school zone sign,white hand,chocolate sprinkled donut,shot glasses,funny faced character,Israeli flag,Japanese flag,Disney cast members,red pouch,green commuter train,fed,rainbow image,green tow truck,ivory tusk,}
  ]
  \addplot coordinates {
    (abandoned, 1)
    (delapidated, 1)
    (light colored countertop, 1)
    (asphalt lot, 1)
    (back of a refrigerator, 1)
    (plastic portable computer, 1)
    (styrofoam tray, 1)
    (right hip, 1)
    (group of three boys, 1)
    (flag bearing motorcycles, 1)
    (Clive, 1)
    (funky looking purple scissors, 1)
    (but, 1)
    (white palm pilot, 1)
    (chicken burger, 1)
    (plug, 1)
    (vandalized school zone sign, 1)
    (white hand, 1)
    (chocolate sprinkled donut, 1)
    (shot glasses, 1)
    (funny faced character, 1)
    (Israeli flag, 1)
    (Japanese flag, 1)
    (Disney cast members, 1)
    (red pouch, 1)
    (green commuter train, 1)
    (fed, 1)
    (rainbow image, 1)
    (green tow truck, 1)
    (ivory tusk, 1)
  };
  \end{axis}
\end{tikzpicture}
\begin{tikzpicture}
\tiny
  \begin{axis}[
    xtick=data,
    ybar,
    xlabel=Least 30 Relation for COCO,
    ylabel=Word frequency,
    width=\textwidth,
    height=6cm,
    enlarge x limits=0.025,
    enlarge y limits=0.2,
    xticklabel style={rotate=90},
    xlabel style={at={(0.5,1.05)}, anchor=south},
    symbolic x coords={either landing or preparing for take off,prepares to land on,side,feeding herself,sailing close to,being mixed,trying to draw water with,feeling for,dangling in front of,setup against,on use by,one hand on the bottom of,waking with,perched high up in,pouring out onto,working near,defending,care for,wrapped up,jumps up after,finished,gather by,brown,flying close above,drinking milk from,reaching head into,squashing,walking single file in,converse with,at each other,}
  ]
  \addplot coordinates {
    (either landing or preparing for take off, 1)
    (prepares to land on, 1)
    (side, 1)
    (feeding herself, 1)
    (sailing close to, 1)
    (being mixed, 1)
    (trying to draw water with, 1)
    (feeling for, 1)
    (dangling in front of, 1)
    (setup against, 1)
    (on use by, 1)
    (one hand on the bottom of, 1)
    (waking with, 1)
    (perched high up in, 1)
    (pouring out onto, 1)
    (working near, 1)
    (defending, 1)
    (care for, 1)
    (wrapped up, 1)
    (jumps up after, 1)
    (finished, 1)
    (gather by, 1)
    (brown, 1)
    (flying close above, 1)
    (drinking milk from, 1)
    (reaching head into, 1)
    (squashing, 1)
    (walking single file in, 1)
    (converse with, 1)
    (at each other, 1)
  };
  \end{axis}
\end{tikzpicture}
 \caption{Plot of the least common entities and relations that were extracted by our LLM-based parser for the COCO datasets.}
  \label{fig:coco_least_common}
\end{figure}

\begin{figure}[ht]
  \centering
\begin{tikzpicture}
\tiny
  \begin{axis}[
    xtick=data,
    ybar,
    xlabel=Top 30 Entities for CC12M,
    ylabel=Word frequency,
    width=\textwidth,
    height=6cm,
    enlarge x limits=0.025,
    enlarge y limits=0.2,
    xticklabel style={rotate=90},
    xlabel style={at={(0.5,1.05)}, anchor=south},
    symbolic x coords={<PERSON>,<person>,person,building,room,white background,bed,woman,hotel,man,beach,wedding,girl,vector illustration,house,photo,people,royalty free illustration,kitchen,bathroom,water,portrait,home,London,table,poster,garden,car,flowers,dog,}
  ]
  \addplot coordinates {
    (<PERSON>, 1105209)
    (<person>, 290129)
    (person, 184634)
    (building, 129756)
    (room, 125712)
    (white background, 116965)
    (bed, 98864)
    (woman, 72515)
    (hotel, 69235)
    (man, 64484)
    (beach, 58818)
    (wedding, 55548)
    (girl, 47695)
    (vector illustration, 42616)
    (house, 42504)
    (photo, 42161)
    (people, 40436)
    (royalty free illustration, 38087)
    (kitchen, 37300)
    (bathroom, 37020)
    (water, 36492)
    (portrait, 36375)
    (home, 36222)
    (London, 33176)
    (table, 33110)
    (poster, 32790)
    (garden, 32373)
    (car, 32165)
    (flowers, 29948)
    (dog, 29805)
  };
  \end{axis}
\end{tikzpicture}
\begin{tikzpicture}
\tiny
  \begin{axis}[
    xtick=data,
    ybar,
    xlabel=Top 30 Relation for CC12M,
    ylabel=Word frequency,
    width=\textwidth,
    height=6cm,
    enlarge x limits=0.025,
    enlarge y limits=0.2,
    xticklabel style={rotate=90},
    xlabel style={at={(0.5,1.05)}, anchor=south},
    symbolic x coords={with,in,on,at,of,for,and,located in,from,by,wearing,has,related to,includes,holding,including,part of,near,featuring,during,to,is,over,in front of,next to,features,using,showing,against,sitting on,}
  ]
  \addplot coordinates {
    (with, 1073543)
    (in, 981344)
    (on, 658344)
    (at, 453675)
    (of, 329627)
    (for, 317847)
    (and, 214960)
    (located in, 179057)
    (from, 144450)
    (by, 137520)
    (wearing, 134435)
    (has, 123377)
    (related to, 122067)
    (includes, 91638)
    (holding, 74895)
    (including, 72025)
    (part of, 59609)
    (near, 59361)
    (featuring, 57766)
    (during, 55396)
    (to, 54933)
    (is, 49526)
    (over, 45001)
    (in front of, 42612)
    (next to, 40557)
    (features, 39894)
    (using, 35223)
    (showing, 34862)
    (against, 34621)
    (sitting on, 33226)
  };
  \end{axis}
\end{tikzpicture}
 \caption{Plot of the most common entities and relations that were extracted by our LLM-based parser for the CC12M dataset.}
  \label{fig:cc12m_most_common}
\end{figure}

\ifarxiv
\begin{figure}
        \centering
        \includegraphics[width=0.5\linewidth]{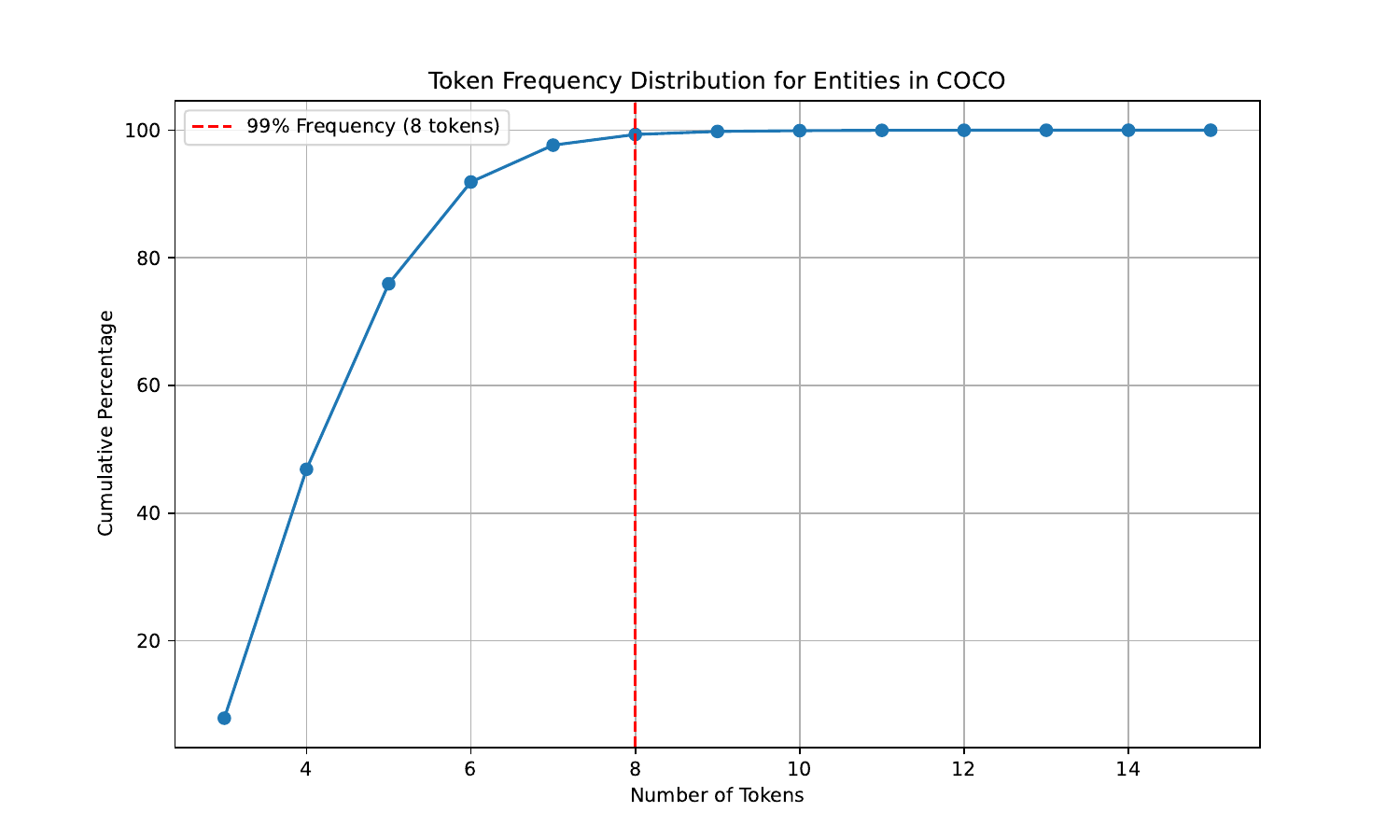}
        \includegraphics[width=0.5\linewidth]{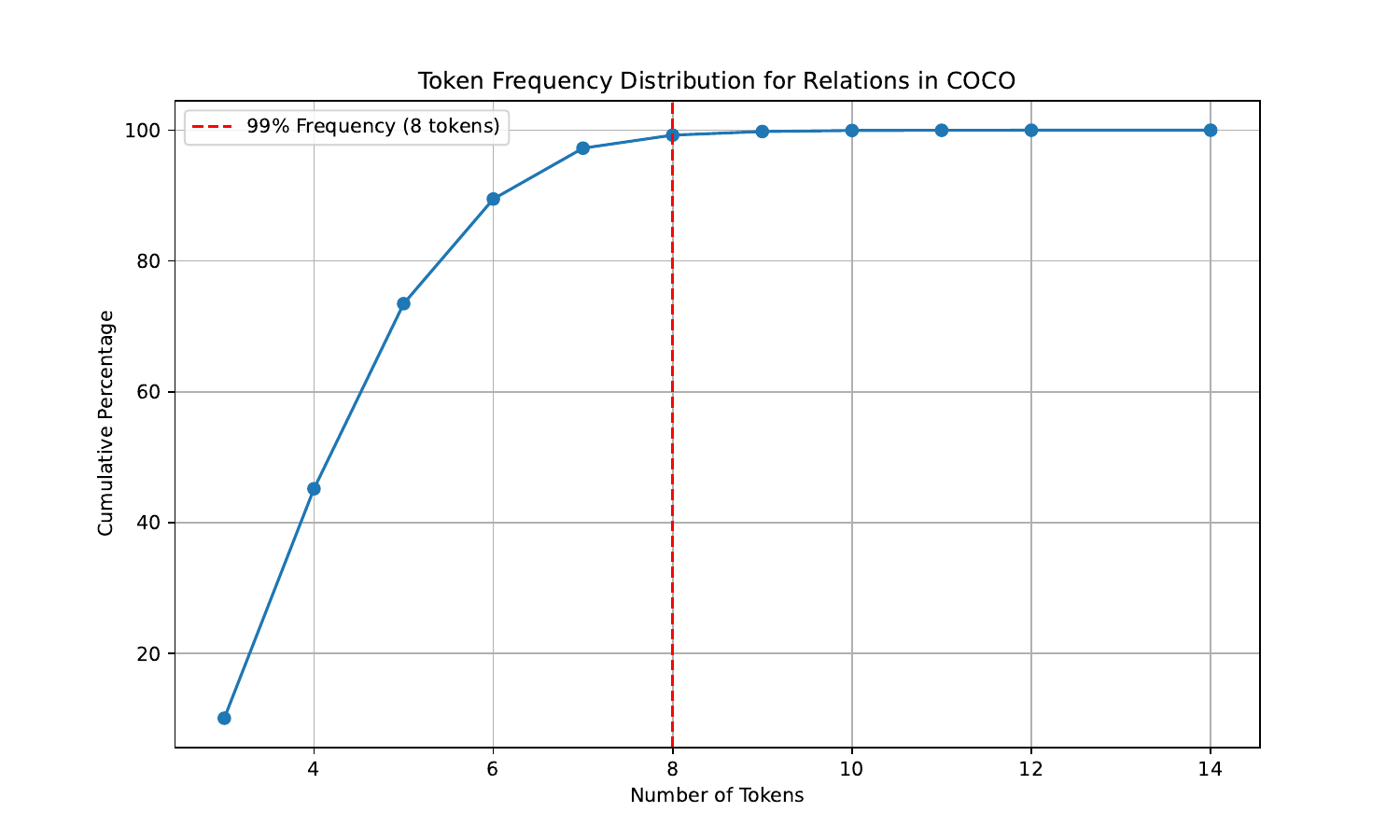}
     \caption{Distribution of the number of tokens require for modeling the entities and relations on COCO (we do not need more than 8 tokens to capture 99\% of the entities in COCO). Since we need less token, we can leverage a smaller text encoder to extract the entities and relations.}
      \label{fig:coco_token_dist}
    \end{figure}

    \begin{figure}
        \centering
        \includegraphics[width=0.5\linewidth]{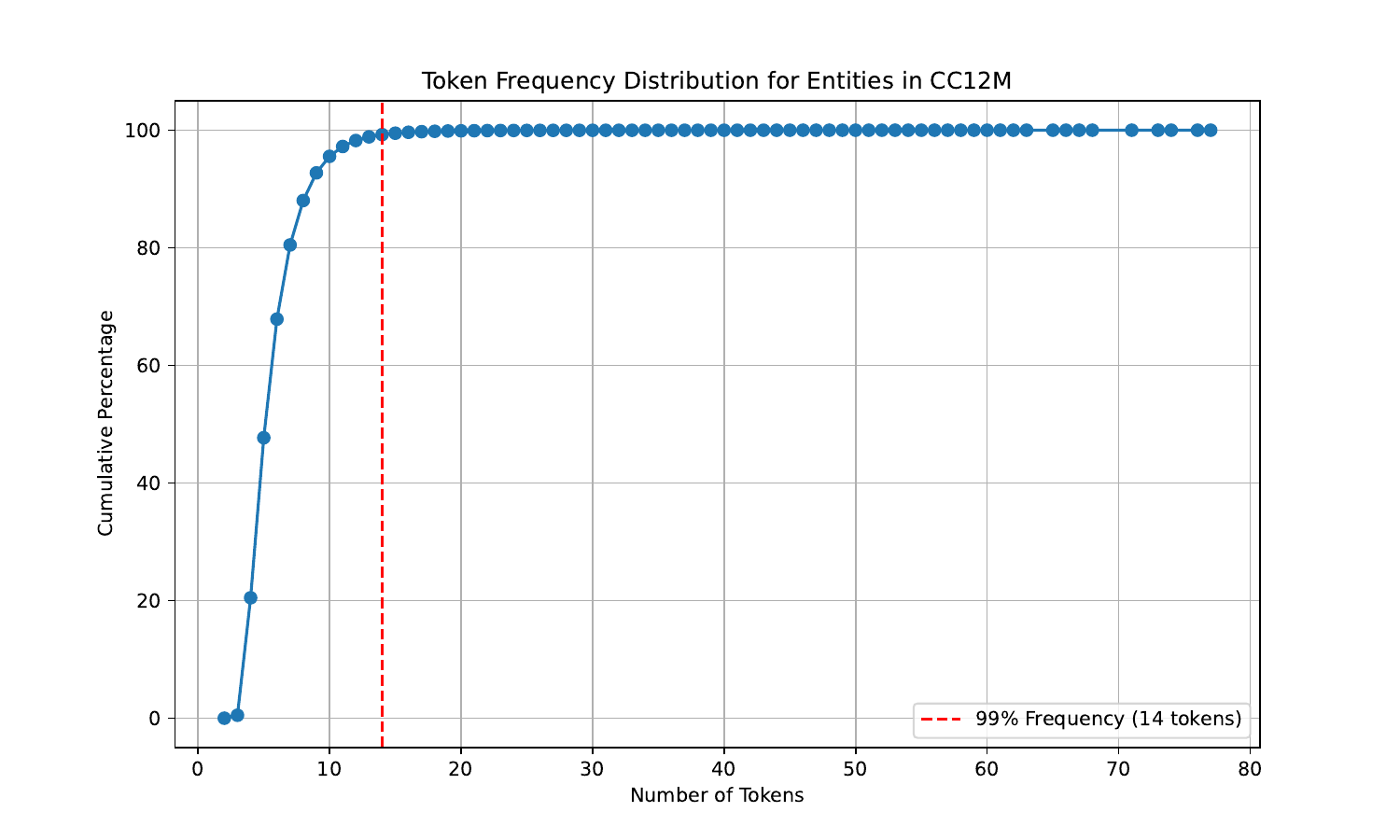}
        \includegraphics[width=0.5\linewidth]{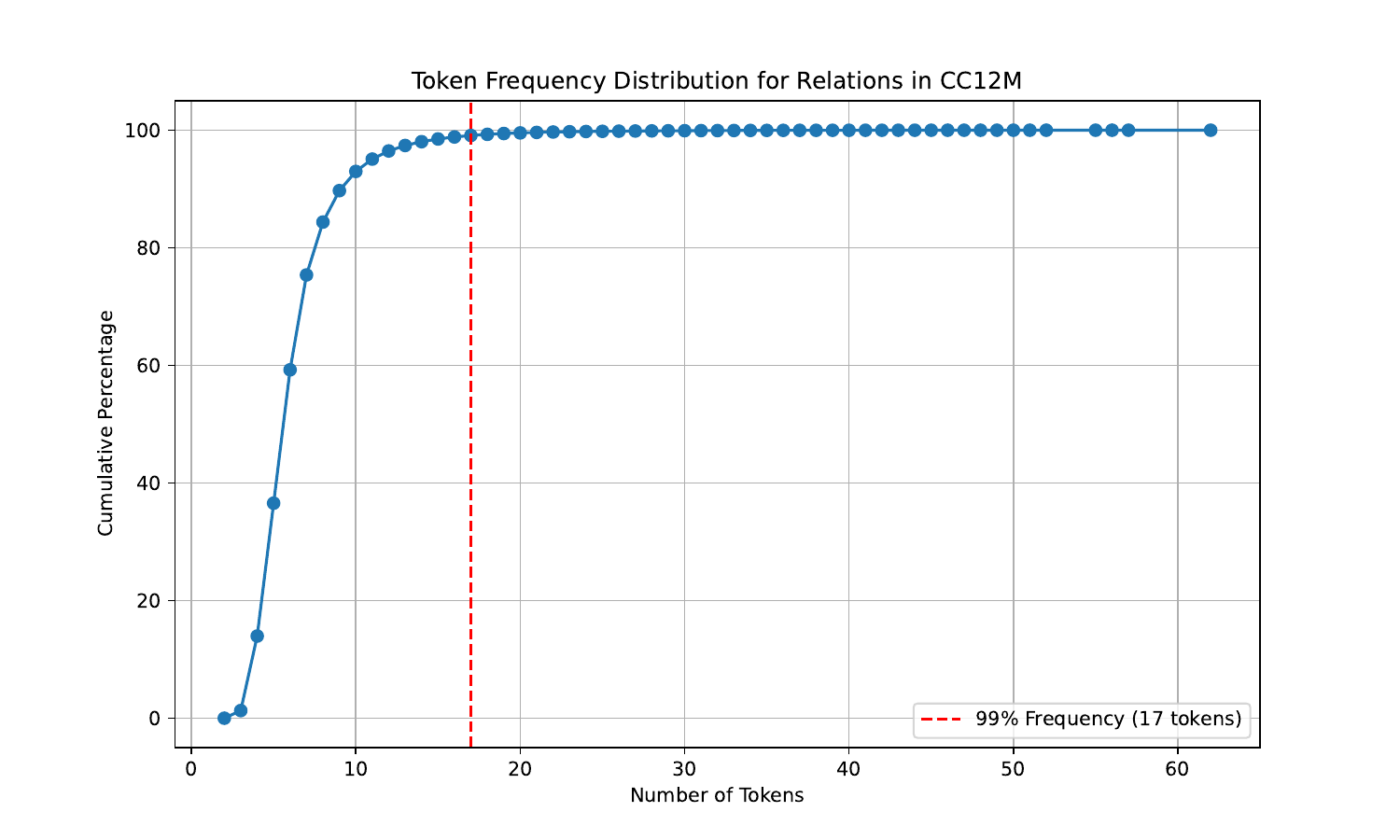}
     \caption{Distribution of the number of tokens require for modeling the entities and relations on CC12M (we do not need more than 14 tokens to capture 99\% of the entities in CC12M). Since we need less token, we can leverage a smaller text encoder to extract the entities and relations.}
      \label{fig:cc12m_token_dist}
    \end{figure}
    
\fi

\subsection{Datasets} 
\paragraph{Training Data}
For the compositional experiments we train both OpenCLIP and OC-CLIP on a aggregated data form COCO-Captions (COCO) \citep{coco}, Visual Genome (VG) \citep{krishna2017visual} and GQA \citep{hudson2019gqa}. All these datasets cover the same 110k images from COCO but focus on different kind of annotations. COCO provide global scene annotation, Visual Genome emphasizes specific region descriptions and general relationships and GQA annotates both objects and spatial relationships. Both Visual Genome and GQA have annotated scene graph that we do not need to parse to train OC-CLIP. For OpenCLIP, we sample 2 region annotations from VG to from a caption following this template \emph{A photo of a \{Region 1\} and a \{Region 2\}}. Similarly to get the captions from GQA, if there is a relationship we follow \citet{kamath2023whatsupvisionlanguagemodels} and give the model a caption following this template \emph{A photo of \{Subject\} \{Rel\} \{Object\}}. If only objects are mentionned we sample up to 3 objects and give the model a caption following this template \emph{A photo of \{Obj1\},\{Obj2\},\{ Obj3\} }.
\subsection{Training Details and Hyperparameters}

In table \ref{tab:hyperparam} we detail the hyperparameters of the OC-CLIP architecture for results in real-world compositional understanding (section 4.2).
\paragraph{Optimization Details} In order to train OC-CLIP we followed prior work and use Adam Optimizer with $\beta_1$ and $\beta_2$ set to 0.9 and 0.95 and a weight decay of 0.2. We used different learning rate for the pretrained backbones and for our modules that we train from scratch : learning rate of $2e^{-4}$ for the binding and the scoring modules, learning rate of $2e^{-5}$ for the text Transformer backbone, and a smaller rate of $1e^{-6}$ for the ViT backbone. We also used a warmup schedule for both of the text (1k steps) and the vision (5k steps) backbones followed by a cosine decay. We train the model for a total of 100 epochs. 
\begin{table}[ht]
\centering
\begin{tabular}{@{}lccc@{}}
\toprule
Hyperparameter/Parameter Init & Architecture & Value\\ 
\midrule
\textbf{Binding Module} &  &  & \\
-- Image Patches Processing & Linear & 768 $\times$ 256 & \\
-- Self-Attention  \#Layers/\#Heads &  & 2/4 & \\
-- Self-Attention MLP ratio/act &  & 2/nn.GELU & \\
-- Keys $K$, Values $V$ & Linear & 256, 256 & \\
-- Normalization Keys/Values & LayerNorm & 256 & \\
\midrule
\textbf{Grouping Module} &  &  & \\
-- Cross-Attention \#Heads &  & 1 & \\
-- Queries & Linear & 256 \\
-- Normalization Queries & LayerNorm & 256 \\
-- Num Default Tokens $Q_{\text{default}}$ & nn.Param($N_d$,256) & 1\\
\midrule
\textbf{Scoring Functions} &  &  & \\

-- Object Scoring Function & cosine sim  &  & \\
-- Relation Scoring subject $ f_s$ &  MLP(128 + 256, 128) & 2 layers & \\
-- Relation Scoring object $ f_o$ &  MLP(128 + 256, 128) & 2 layers & \\
-- Coef ent init (learned parameter) & &1.5 \\
-- Coef rel init (learned parameter) & & 0.5\\

 \bottomrule
\end{tabular}%
\caption{Table of hyperparameters for OC-CLIP architecture}
\label{tab:hyperparam}
\end{table}

\subsection{Binding Module Code}
\begin{figure}
    \centering
    \includegraphics[width=\linewidth]{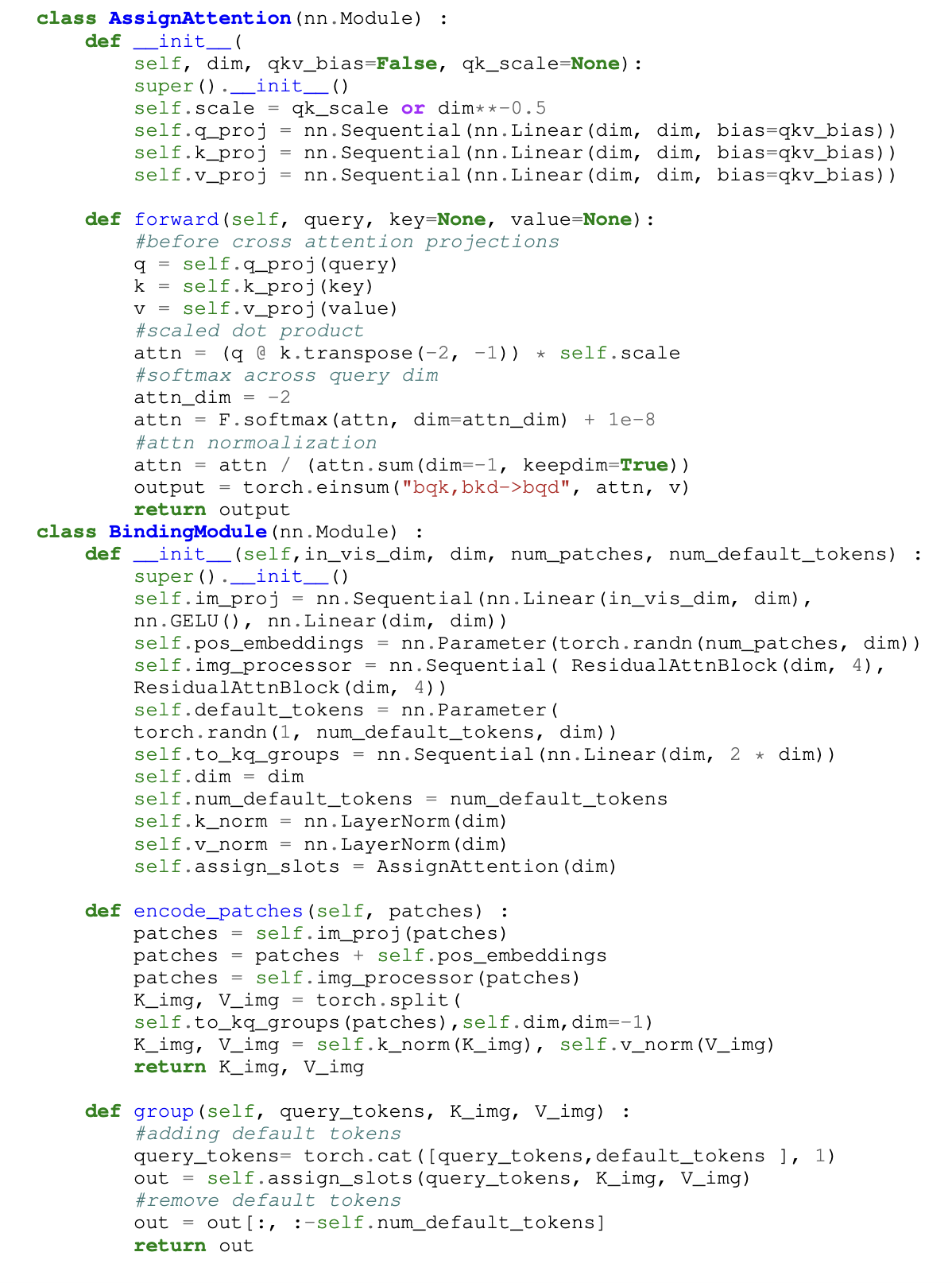}
    \caption{Code for the Binding Module}
    \label{fig:code}
\end{figure}

See Figure \ref{fig:code}

\begin{figure}[ht]
  \centering
    \begin{tikzpicture}
    \tiny
      \begin{axis}[
        xtick=data,
        ybar,
        xlabel=Top 30 Entities for CC3M,
        ylabel=Word Frequency,
        width=\textwidth,
        height=6cm,
        enlarge x limits=0.025,
        enlarge y limits=0.2,
        xticklabel style={rotate=90},
        xlabel style={at={(0.5,1.05)}, anchor=south},
        symbolic x coords={person,actor,white background,city,man,premiere,beach,woman,people,view,football player,stage,festival,portrait,musical instrument,sports team,ball,river,model,water,car,pop artist,girl,vector illustration,house,tree,event,sunset,politician,flag,}
      ]
      \addplot coordinates {
        (person, 253338)
        (actor, 109367)
        (white background, 55313)
        (city, 46557)
        (man, 40357)
        (premiere, 39363)
        (beach, 37933)
        (woman, 33204)
        (people, 32698)
        (view, 31104)
        (football player, 30497)
        (stage, 30222)
        (festival, 29221)
        (portrait, 26613)
        (musical instrument, 25862)
        (sports team, 25470)
        (ball, 23177)
        (river, 21091)
        (model, 20986)
        (water, 18405)
        (car, 18364)
        (pop artist, 18161)
        (girl, 18079)
        (vector illustration, 17925)
        (house, 17882)
        (tree, 17764)
        (event, 17721)
        (sunset, 17632)
        (politician, 16641)
        (flag, 16300)
      };
    \end{axis}
    \end{tikzpicture}
    \begin{tikzpicture}
    \tiny
      \begin{axis}[
        xtick=data,
        ybar,
        xlabel=Top 30 Relation for CC3M,
        ylabel=Word Frequency,
        width=\textwidth,
        height=6cm,
        enlarge x limits=0.025,
        enlarge y limits=0.2,
        xticklabel style={rotate=90},
        xlabel style={at={(0.5,1.05)}, anchor=south},
        symbolic x coords={on,with,in,at,during,of,attends,against,from,over,for,by,wearing,and,attend,on stage,holding,arrives at,playing,in front of,performs on,sitting on,near,under,isolated on,to,walks on,along,behind,looking at,}
      ]
      \addplot coordinates {
        (on, 249858)
        (with, 216254)
        (in, 205332)
        (at, 85210)
        (during, 64058)
        (of, 45280)
        (attends, 43674)
        (against, 35350)
        (from, 30801)
        (over, 27717)
        (for, 26672)
        (by, 25561)
        (wearing, 25397)
        (and, 22231)
        (attend, 22125)
        (on stage, 21332)
        (holding, 21096)
        (arrives at, 19263)
        (playing, 17608)
        (in front of, 17030)
        (performs on, 16946)
        (sitting on, 16848)
        (near, 14713)
        (under, 14603)
        (isolated on, 14316)
        (to, 13172)
        (walks on, 12068)
        (along, 10332)
        (behind, 9357)
        (looking at, 9337)
      };
    \end{axis}
    \end{tikzpicture}
 \caption{Top entities and relation for CC3M}
\end{figure}

\begin{figure}[ht]
  \centering
\begin{tikzpicture}
\tiny
  \begin{axis}[
    xtick=data,
    ybar,
    xlabel=Top 30 Entities for CC12M,
    ylabel=Word Frequency,
    width=\textwidth,
    height=6cm,
    enlarge x limits=0.025,
    enlarge y limits=0.2,
    xticklabel style={rotate=90},
    xlabel style={at={(0.5,1.05)}, anchor=south},
    symbolic x coords={<PERSON>,<person>,person,building,room,white background,bed,woman,hotel,man,beach,wedding,girl,vector illustration,house,photo,people,royalty free illustration,kitchen,bathroom,water,portrait,home,London,table,poster,garden,car,flowers,dog,}
  ]
  \addplot coordinates {
    (<PERSON>, 1105209)
    (<person>, 290129)
    (person, 184634)
    (building, 129756)
    (room, 125712)
    (white background, 116965)
    (bed, 98864)
    (woman, 72515)
    (hotel, 69235)
    (man, 64484)
    (beach, 58818)
    (wedding, 55548)
    (girl, 47695)
    (vector illustration, 42616)
    (house, 42504)
    (photo, 42161)
    (people, 40436)
    (royalty free illustration, 38087)
    (kitchen, 37300)
    (bathroom, 37020)
    (water, 36492)
    (portrait, 36375)
    (home, 36222)
    (London, 33176)
    (table, 33110)
    (poster, 32790)
    (garden, 32373)
    (car, 32165)
    (flowers, 29948)
    (dog, 29805)
  };
  \end{axis}
\end{tikzpicture}
\begin{tikzpicture}
\tiny
  \begin{axis}[
    xtick=data,
    ybar,
    xlabel=Top 30 Relation for CC12M,
    ylabel=Word Frequency,
    width=\textwidth,
    height=6cm,
    enlarge x limits=0.025,
    enlarge y limits=0.2,
    xticklabel style={rotate=90},
    xlabel style={at={(0.5,1.05)}, anchor=south},
    symbolic x coords={with,in,on,at,of,for,and,located in,from,by,wearing,has,related to,includes,holding,including,part of,near,featuring,during,to,is,over,in front of,next to,features,using,showing,against,sitting on,}
  ]
  \addplot coordinates {
    (with, 1073543)
    (in, 981344)
    (on, 658344)
    (at, 453675)
    (of, 329627)
    (for, 317847)
    (and, 214960)
    (located in, 179057)
    (from, 144450)
    (by, 137520)
    (wearing, 134435)
    (has, 123377)
    (related to, 122067)
    (includes, 91638)
    (holding, 74895)
    (including, 72025)
    (part of, 59609)
    (near, 59361)
    (featuring, 57766)
    (during, 55396)
    (to, 54933)
    (is, 49526)
    (over, 45001)
    (in front of, 42612)
    (next to, 40557)
    (features, 39894)
    (using, 35223)
    (showing, 34862)
    (against, 34621)
    (sitting on, 33226)
  };
  \end{axis}
\end{tikzpicture}
 \caption{Top entities and relation for CC12M}
\end{figure}

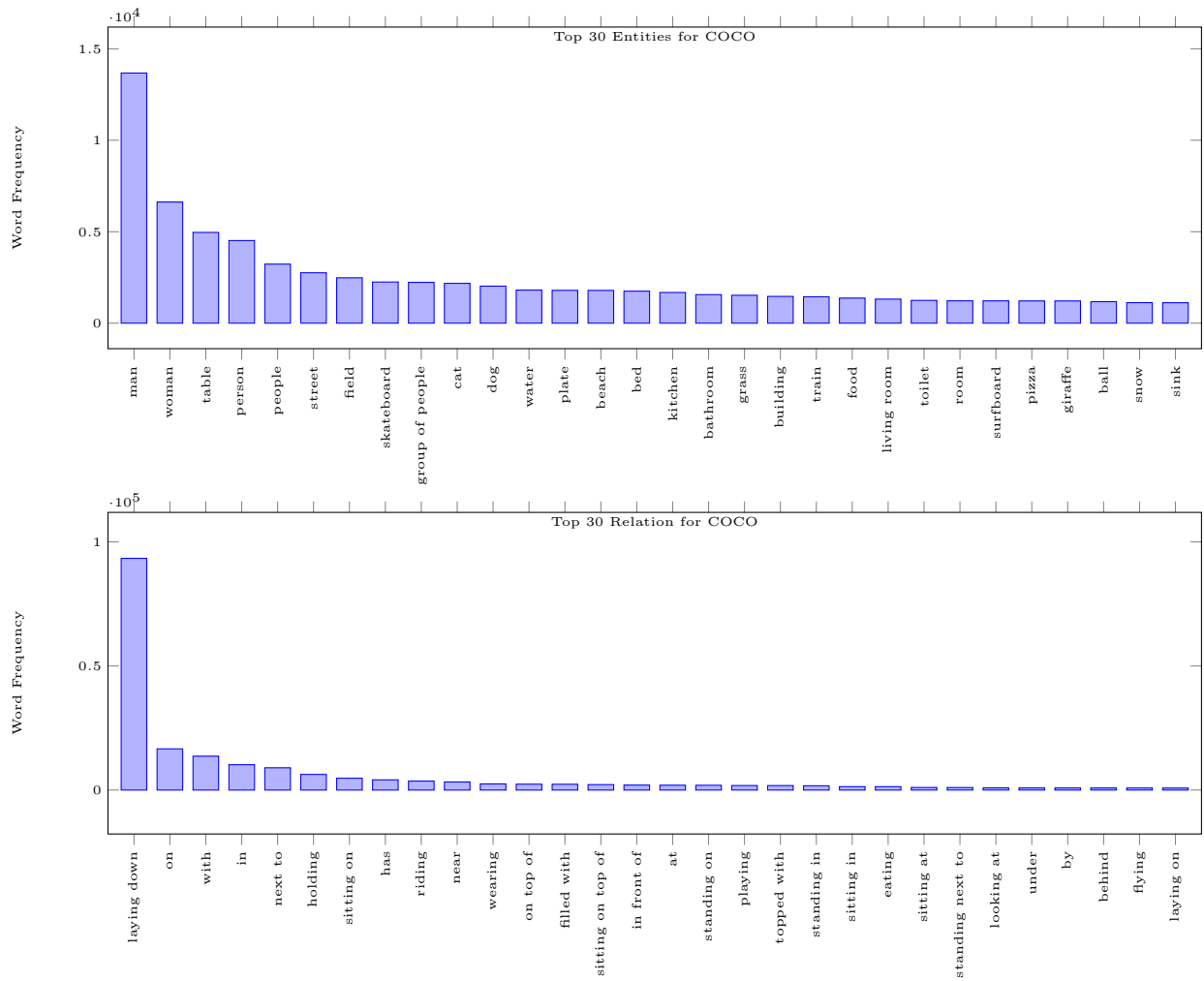
\begin{figure}[ht]
  \centering
\begin{tikzpicture}
\tiny
  \begin{axis}[
    xtick=data,
    ybar,
    xlabel=Top 30 Entities for COCO,
    ylabel=Word Frequency,
    width=\textwidth,
    height=6cm,
    enlarge x limits=0.025,
    enlarge y limits=0.2,
    xticklabel style={rotate=90},
    xlabel style={at={(0.5,1.05)}, anchor=south},
    symbolic x coords={man,woman,table,person,people,street,field,skateboard,group of people,cat,dog,water,plate,beach,bed,kitchen,bathroom,grass,building,train,food,living room,toilet,room,surfboard,pizza,giraffe,ball,snow,sink,}
  ]
  \addplot coordinates {
    (man, 13676)
    (woman, 6624)
    (table, 4963)
    (person, 4518)
    (people, 3228)
    (street, 2756)
    (field, 2482)
    (skateboard, 2251)
    (group of people, 2233)
    (cat, 2178)
    (dog, 2025)
    (water, 1809)
    (plate, 1795)
    (beach, 1792)
    (bed, 1750)
    (kitchen, 1679)
    (bathroom, 1563)
    (grass, 1526)
    (building, 1463)
    (train, 1444)
    (food, 1372)
    (living room, 1315)
    (toilet, 1247)
    (room, 1222)
    (surfboard, 1221)
    (pizza, 1216)
    (giraffe, 1215)
    (ball, 1176)
    (snow, 1123)
    (sink, 1118)
  };
  \end{axis}
\end{tikzpicture}
\begin{tikzpicture}
\tiny
  \begin{axis}[
    xtick=data,
    ybar,
    xlabel=Top 30 Relation for COCO,
    ylabel=Word Frequency,
    width=\textwidth,
    height=6cm,
    enlarge x limits=0.025,
    enlarge y limits=0.2,
    xticklabel style={rotate=90},
    xlabel style={at={(0.5,1.05)}, anchor=south},
    symbolic x coords={laying down,on,with,in,next to,holding,sitting on,has,riding,near,wearing,on top of,filled with,sitting on top of,in front of,at,standing on,playing,topped with,standing in,sitting in,eating,sitting at,standing next to,looking at,under,by,behind,flying,laying on,}
  ]
  \addplot coordinates {
    (laying down, 93385)
    (on, 16607)
    (with, 13665)
    (in, 10224)
    (next to, 8983)
    (holding, 6307)
    (sitting on, 4772)
    (has, 4126)
    (riding, 3618)
    (near, 3254)
    (wearing, 2474)
    (on top of, 2374)
    (filled with, 2344)
    (sitting on top of, 2206)
    (in front of, 2007)
    (at, 1962)
    (standing on, 1932)
    (playing, 1832)
    (topped with, 1820)
    (standing in, 1740)
    (sitting in, 1391)
    (eating, 1390)
    (sitting at, 1052)
    (standing next to, 1031)
    (looking at, 890)
    (under, 860)
    (by, 855)
    (behind, 849)
    (flying, 831)
    (laying on, 817)
  };
 \end{axis}
\end{tikzpicture}
 \caption{Top entities and relation for COCO}
\end{figure}

\end{document}